# A Soft Sensor Method with Uncertainty-Awareness and Self-Explanation Based on Large Language Models Enhanced by Domain Knowledge Retrieval

## Author Information

Shuo Tong, Han Liu ✉, Runyuan Guo, Wenqing Wang, Xueqiong Tian, Lingyun Wei, Lin Zhang, Huayong Wu, Ding Liu, Youmin Zhang

✉ e-mail: liuhan@xaut.edu.cn

## Affiliations

**School of Automation and Information Engineering, Xi'an University of Technology, Xi'an, China.**

Shuo Tong, Han Liu, Runyuan Guo, Wenqing Wang, Xueqiong Tian, Lingyun Wei, Lin Zhang, Huayong Wu, Ding Liu

**Department of Mechanical, Industrial, and Aerospace Engineering and the Concordia Institute of Aerospace Design and Innovation, Concordia University, Montreal, Canada.**

Youmin Zhang

## Corresponding author

Correspondence to: Han Liu

## Abstract

Data-driven soft sensors are crucial in predicting key performance indicators in industrial systems. However, current methods predominantly rely on the supervised learning paradigms of parameter updating, which inherently faces challenges such as high development costs, poor robustness, training instability, and lack of interpretability. Recently, large language models (LLMs) have demonstrated significant potential across various domains, notably through In-Context Learning (ICL), which enables high-performance task execution with minimal input-label demonstrations and no prior training. This paper aims to replace supervised learning with the emerging ICL paradigm for soft sensor modeling to address existing challenges and explore new avenues for advancement. To achieve this, we propose a novel framework called the Few-shot Uncertainty-aware and self-Explaining Soft Sensor (LLM-FUESS), which includes the LLM-based Zero-shot Auxiliary Variable Selector (LLM-ZAVS) and the LLM-based Uncertainty-aware Few-shot Soft Sensor (LLM-UFSS). The LLM-ZAVS retrieves from the Industrial Knowledge Vector Storage (IKVS) to enhance LLMs' domain-specific knowledge, enabling zero-shot auxiliary variable selection. In the LLM-UFSS, we utilize text-based context demonstrations of structured data to prompt LLMs to execute ICL for predicting and propose a context sample retrieval augmentation strategy to improve performance. Additionally, we explored LLMs' AI-Generated Content (AIGC) and probabilistic characteristics to propose self-explanation and uncertainty quantification methods for constructing a trustworthy soft sensor. Extensive experiments on industrial datasets from bioprocessing and chemical engineering demonstrate that our method achieved state-of-the-art predictive performance, strong robustness, and flexibility, effectively mitigates training instability found in traditional methods. To the best of our knowledge, this is the first work to establish soft sensor utilizing LLMs.

## Introduction

In modern industrial processes, accurate monitoring of key quality variables is an indispensable requirement for ensuring the safe and efficient operation of automation systems[1]. Unfortunately, direct online measurement of certain critical variables is often impractical due to high equipment costs, harsh environments, and technological constraints[2,3]. As a result, soft sensor technology has emerged as an efficient solution, utilizing readily accessible process variables (auxiliary variables) to model and predict difficult-to-measure quality variables (primary variables) in real-time. This technology plays an increasingly vital role in industrial process optimization, control, and product quality monitoring[4].

Soft sensor modeling methods can be primarily classified into mechanism-based and data-driven approaches[5]. Data-driven models rely on historical data from distributed control systems (DCS), bypassing the intricacies of process mechanisms, which has made them the mainstream approach today[6,7]. Traditional data-driven soft sensors typically include principal component analysis (PCA) [8,9], partial least squares regression (PLSR) [10,11], and support vector machines (SVM) [12], along with their variants. Recently, deep learning has advanced rapidly, garnering significant attention for its powerful capabilities in handling nonlinear relationships and extracting complex features. Popular deep learning-based soft sensing techniques include multilayer perceptrons (MLP) [13], Long Short-Term Memory networks (LSTM) [14], and Stacked Autoencoders (SAE) [15,16].

Traditional statistical methods, machine learning-based soft sensors, and emerging deep learning soft sensors (collectively referred to as numerical models) all operates on the paradigm of supervised learning. This involves optimizing an objective function, whereby the model iteratively adjusts its parameters on the training dataset to capture complex input-output relationships. However, these numerical models face inherent limitations that significantly hinder their practical application, which we categorize into four main aspects: **(1) High modeling costs and barriers to entry:** Developing soft sensors often requires task-specific design and training tailored to specific on-site conditions. This fragmented and customized approach greatly increases the time and computational costs, along with the complexity of development and maintenance. Moreover, building complex models is knowledge-intensive, necessitating data analysts with expertise across various data science and algorithmic fields, creating a high entry barrier for soft sensor modeling. **(2) Limited robustness and flexibility with input data:** Numerical models impose stringent requirements on the format, dimensions, quantity, and types of input data[17,18]. Raw data must be normalization and dimensional unification, complicating preprocessing. Additionally, these models cannot handle missing values, which restricts their flexibility[19,20]. Finally, they rely solely on structured data from industrial processes, making it difficult to incorporate relevant mechanistic knowledge and context, thus limiting their representation learning potential. **(3) Instability during the training process:** Numerical models are sensitive to initial parameters, sample sizes, and hyperparameter settings[21]. An excessive or insufficient number of samples can lead to overfitting or underfitting, while random fluctuations and changes in data distribution during training can result in issues such as gradient explosion or vanishing gradients[22]. **(4) Lack of Interpretability and Uncertainty Quantification:** Numerical models feature complex structures involving multiple layers of nonlinear relationships and numerous parameters, making it hard to interpret the connections between input features and outcomes[23,24]. Furthermore, they identify hidden patterns in a data-driven manner, lacking transparency based on rules or explicitly theories, which complicates the provision of clear decision-making support for prediction. Moreover, these models rely on deterministic algorithms that often yield single-point predictions without quantifying uncertainty in input data or the model itself. This inability to reflect potential risks or error ranges remains an unresolved issues in the field of soft sensor[25].

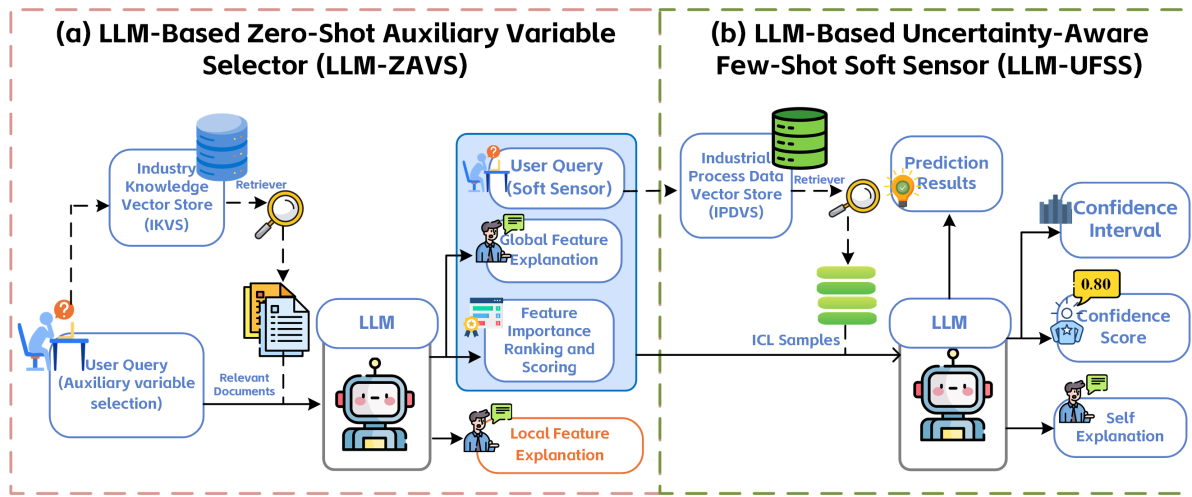

**Fig. 1 | Overview of the design of the proposed LLM-FUESS.** This two-stage few-shot soft sensing framework consists of LLM-ZAVS, which performs auxiliary variable selection, and LLM-UFSS, which executes the soft sensing tasks. The entire pipeline is relies on natural language text input, utilizing ICL, retrieval-augmented generation, and advanced prompting techniques to achieve accurate predictions and generate detailed AIGC self-explanations, all without model training or parameter updates.

Recently, large language models (LLMs) pretrained on extensive corpora has provided promising solutions to these challenges. Models like GPT-4[26] and Gemini-1.5-pro[27] have garnered considerable attention natural language processing (NLP) for their impressive text generation and reasoning abilities[28]. They have also achieved remarkable success in complex domains beyond NLP, including dermatological diagnosis[29], mathematical reasoning[30], patient records interpretation[31], and chemistry[32], continuously pushing the boundaries of LLMs capabilities. This success is largely attributed to their emergent abilities, which develop new advanced capabilities as model parameters scale up[33,34]. A key ability is In-Context

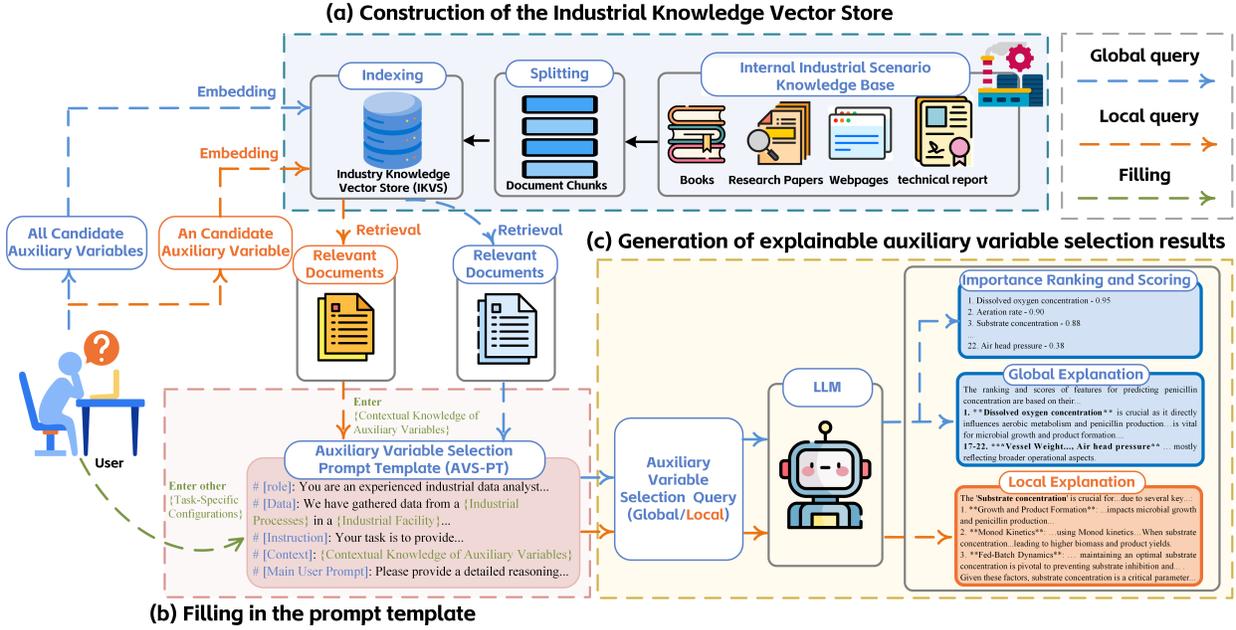

**Fig. 2 | Overview of LLM-ZAVS for zero-shot auxiliary variable selection.** This module involves three steps: (a) constructing an industrial knowledge vector store, (b) populating prompt templates, and (c) generating explainable auxiliary results using LLMs. Two query methods are designed to address different task requirements: the global query provides an importance ranking of all variables along with a comprehensive explanation, while the local query evaluates the impact of a specific variable on the primary variable for localized explanation.

Learning (ICL) [35,36], allowing LLMs to rapidly adapt to unseen tasks by learning from a few examples (input-output pairs) in prompts, and to return results without any additional training or parameter updates.

Encouraged by this, this paper aims to leverage the new learning paradigm of ICL to replace the traditional supervised learning pipeline for soft sensor modeling, addressing challenges in data-driven soft sensor without sacrificing predictive performance. As illustrated in Fig. 1, we propose a few-shot soft sensor framework based on LLMs, called LLM-FUESS, which incorporates uncertainty-awareness and self-explanation. LLM-FUESS streamline the soft sensor pipeline into two stages: auxiliary variable selection and soft sensor modeling, named the LLM-Based Zero-Shot Auxiliary Variable Selector (LLM-ZAVS) and the LLM-Based Uncertainty-Aware Few-Shot Soft Sensor (LLM-UFSS), respectively.

Selecting auxiliary variables is essential in soft sensor modeling, as it minimizes interference from irrelevant variables and reduces prompt token counts[37], lowering API costs. As depicted in Fig. 2, we propose a zero-shot auxiliary variable selector (LLM-ZAVS) that uses LLMs and prompt learning to generate reasonable feature selection results, along with importance rankings and scores, without examining any data samples. While pretrained LLMs encode extensive knowledge, they may lack detailed, domain-specific insights about industrial processes, material properties, and reaction mechanisms, leading to "hallucinations" and unreliable feature selection. To address this, we integrate domain-specific industrial knowledge with LLMs using Retrieval-Augmented Generation (RAG)[38,39], transforming LLMs into informed domain experts for selecting auxiliary variables. To ensure the reliability and transparency of LLMs decisions, we introduce the Chain of Thought (CoT) [40,41], enabling LLMs to provide detailed explanations of the decision-making process (both globally and locally). Additionally, we design a new evaluation metric called the Average Selection Consistency Score (ASCS) to assess the consistency of LLM-ZAVS's results. Experimental reults shows that LLM-ZAVS performs competitive against other numerical feature selection methods and exhibits high output consistency.

In the LLM-UFSS (Fig. 3), we format structured process data into text-based input-output pairs as context demonstration samples. Unlike supervised methods, LLMs capture hidden nonlinear patterns among variables by learning from few-shot context examples in prompts, enabling accurate predictions of the target variable for test samples. Since the entire framework relies on prompt strategies without any modifications to model or training, **LLM-UFSS is both code-free and model-free, significantly reducing modeling complexity, time costs, and dependency on specialized knowledge.** Given that the performance of ICL depends on the quality of demonstration samples[42], we propose a context sample enhancement strategy inspired by RAG. Unlike traditional RAG, which focuses on document retrieval and augmentation, our method utilizes industrial sample vectors. Specifically, we use the test sample as a query to retrieve similar samples from the constructed Industrial Process Data Vector Store (IPDVS) for improved the quality of context demonstrations. Notably, the input to LLM-UFSS is entirely in a user-friendly natural language format. Due to the inherent robustness of LLMs to prompts, we are able to bypass normalization and imputation for missing values, **demonstrating strong flexibility in handling inputs.** leveraging LLMs' powerful multimodal capabilities, we integrate industrial domain knowledge texts (such as background, data information, and mechanistic knowledge) into the prompts, combining them with numerical data. **This multimodal fusion significantly enriches the model's expression by enhancing information from multiple perspectives.** Furthermore, we instruct LLMs to provide detailed, step-by-step, and human-readable explanations of the decision-making process during prediction, **enhancing the interpretability and transparency of the proposed method.** Finally, we propose two uncertainty quantification methods for soft sensing using LLMs' probabilistic characteristics: constructing confidence intervals and outputting confidence scores, **enhancing the method's credibility, reliability, and risk awareness.** Through extensive quantitative analysis and ablation experiments, we demonstrate the strong performance and capabilities of LLM-FUESS from multiple aspects. Remarkably, our experiments also indicate that **LLM-UFSS can effectively prevent training instability issues such as overfitting or underfitting commonly encountered in supervised learning methods.**

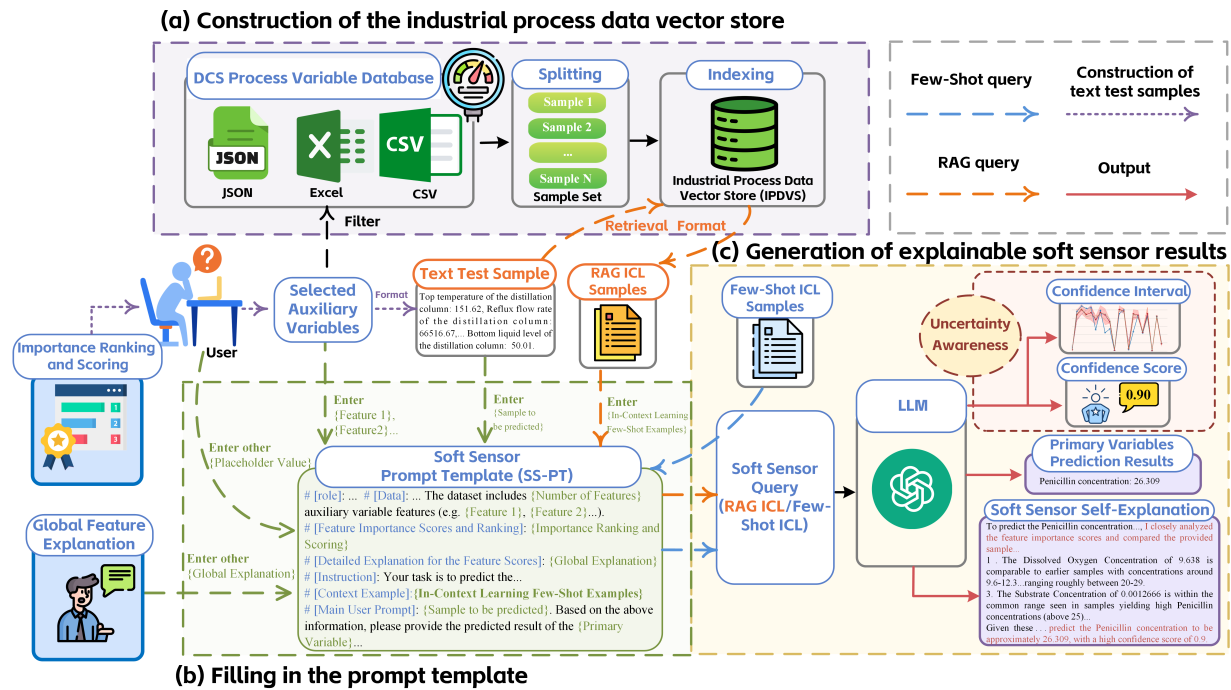

**Fig. 3 | Overview of the LLM-UFSS for few-shot soft sensing.** This module consists of three steps: (a) constructing an industrial process data vector store, (b) populating templates and enhancing context with sample retrieval, and (c) generating explainable soft sensor results with uncertainty awareness.

Finally, we incorporate various prompt engineering techniques such as role-play[43], CoT, and emotional stimulation[44] to create two fixed fill-in-the-blank prompt templates for LLM-ZAVS and LLM-

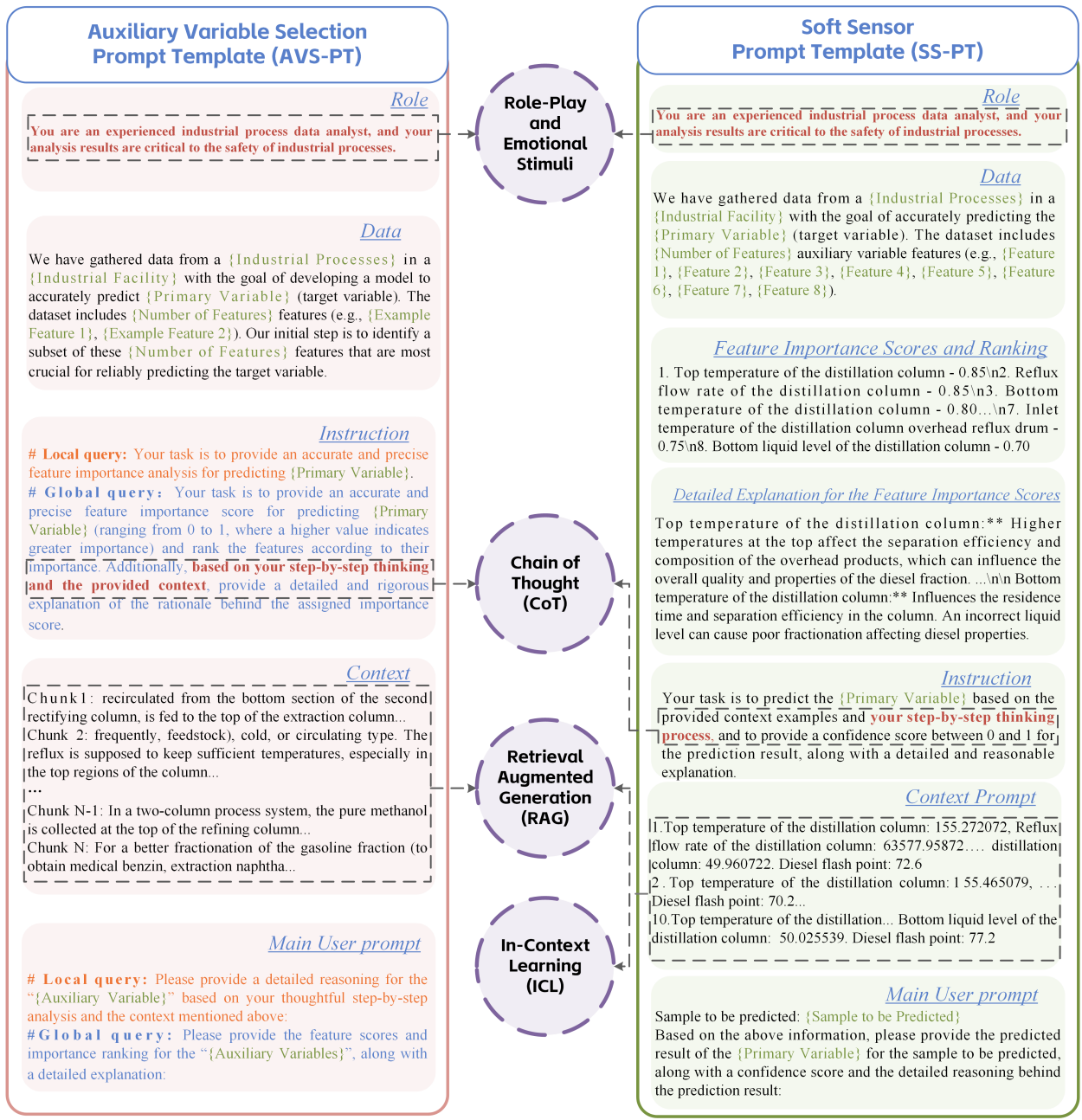

**Fig. 4 | Structure diagrams of AVS-PT and SS-PT.** The AVS-PT consists of five elements: Role, Data, Instruction, Context, and Main User Prompt. The SS-PT additionally includes Feature Importance Scores and Ranking, as well as a Detailed Explanation for the Feature Importance Scores, generated by the LLM-ZAVS.

UFSS: the Auxiliary Variable Selection Prompt Template (AVS-PT) and the Soft Sensor Prompt Template (SS-PT) (Fig. 4). Using LangChain, we encapsulated the entire process framework, allowing users to simply fill in a few external keywords into the templates according to operational requirements.

The framework then automatically executes the corresponding tasks end-to-end, delivering stable results. This approach aligns with human interaction methods, making it user-friendly for non-experts without AI or coding backgrounds. **By changing only a few keywords, the template can adapt to all soft sensing tasks, simplifying manual configuration and operation.**

## Results

### Case study

To validate the proposed method's performance and general applicability in industrial process, we selected the penicillin fermentation process (biocatalytic reactions) and the polypropylene production process (chemical polymerization) as case studies (Fig. 5).

(1) Penicillin fermentation process

Penicillin, a secondary metabolite synthesized by the penicillium mold under specific conditions, is one of the most widely used antibiotics. As shown in Fig. 5a, the fermentation process is a complex nonlinear batch process. However, the absence of reliable sensors for real-time measurement of product concentration, a critical quality variable, makes effective control of the fermentation process challenging. Therefore, developing a soft sensor to measure penicillin concentration in real-time and accurately using easily accessible process variables is of significant research interest.

In this study, the industrial-scale penicillin fermentation simulation (IndPensim)[45] was selected for experimentation. IndPensim integrates the complex characteristics of various industrial-scale processes and serves as a simulation platform closely resembling actual industrial penicillin fed-batch fermentation processes. We selected three normal batches with different parameter settings from IndPensim, with fermentation times of 226h, 230h, and 278h, and a sampling interval of 0.2h. Each batch comprises two phases: the batch phase and the fed-batch phase. In the initial batch phase, Penicillium consumes the substrate, leading to rapid mycelial growth, but penicillin is not produced. After 24 hours, the process transitions to the fed-batch phase, during which a substantial amount of penicillin is continuously synthesized. Our experiments focus on data from fed-batch phase (post-24h), predicting penicillin concentration as the primary variable, while 22 accessible process variables serve as auxiliary variable for feature selection and soft sensor modeling. Detailed descriptions and units are in Table 1.

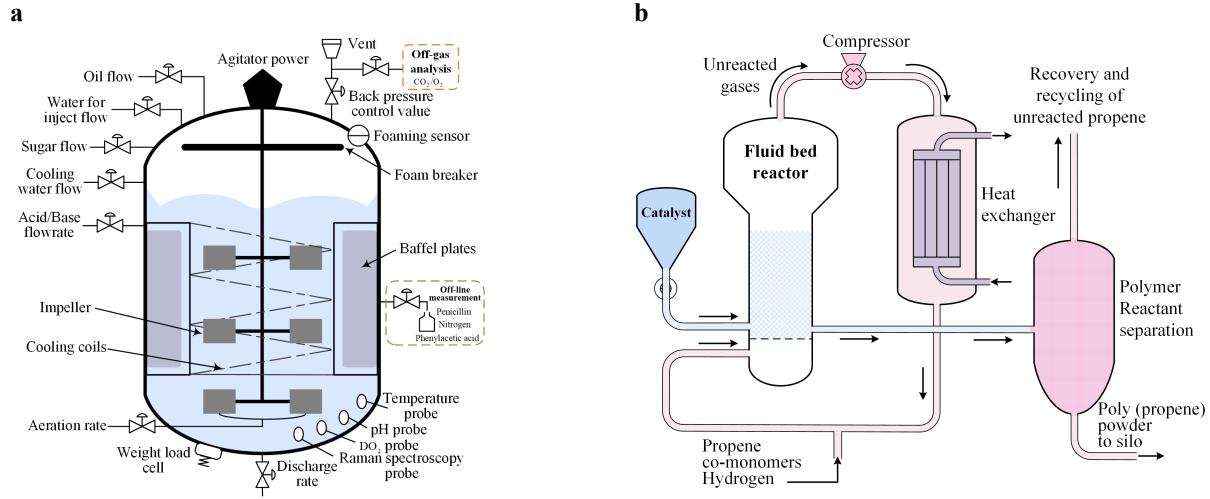

**Fig. 5 | Case study diagrams. a**, Schematic diagram of the penicillin fermentation process. **b**, Schematic diagram of the polypropylene production process.

**Table 1 | Variables of the penicillin fermentation process**

| Variable | Description | Unit | Variable | Description | Unit |
| --- | --- | --- | --- | --- | --- |
| V1 | Aeration rate | $m^3\,min^{-1}$ | V12 | Vessel Volume | L |
| V2 | Sugar feed rate | $Lh^{-1}$ | V13 | Vessel Weight | Kg |
| V3 | Acid flow rate | $Lh^{-1}$ | V14 | pH | / |
| V4 | Base flow rate | $Lh^{-1}$ | V15 | Temperature of broth | K |
| V5 | Heating/cooling water flow rate | $Lh^{-1}$ | V16 | Generated heat | KJ |
| V6 | Heating water flow rate | $Lh^{-1}$ | V17 | carbon dioxide percent in off-gas | % |
| V7 | Water for injection/dilution | $Lh^{-1}$ | V18 | PAA flow | $Lh^{-1}$ |
| V8 | Air head pressure | bar | V19 | Oil flow | $Lh^{-1}$ |
| V9 | Dumped broth flow | $Lh^{-1}$ | V20 | Oxygen Uptake Rate | $g\,min^{-1}$ |
| V10 | Substrate concentration | $gL^{-1}$ | V21 | Oxygen in percent in off-gas | % |
| V11 | Dissolved oxygen concentration | $mgL^{-1}$ | V22 | Carbon evolution rate | $gh^{-1}$ |

(2) Polypropylene production process

Polypropylene is one of the most widely used and lightest thermoplastic polymers, produced through chain-growth polymerization of the monomer propylene using catalysts under controlled temperature and pressure. As depicted in Fig. 5b, the production process begins with the mixing of propylene, comonomers, and chain transfer agents (such as hydrogen) flowing into a fluidized bed reactor containing a high-activity catalyst to produce polypropylene powder. Unreacted gases are recycled back into the reaction system through compressors and heat exchangers. The resulting polypropylene powder is then transported to a cyclone separator to remove gaseous propylene and hydrogen, followed by further processing and molding into various plastic products.

During production, excessively long or short polymer chains can lead to polymer viscosity levels that are too high or too low, failing to meet customer requirements. Measuring the Melt Flow Rate (MFR) is crucial for quality control as it indicates the viscosity of the thermoplastic polymer melt. However, MFR is typically estimated in the laboratory every 2-8 hours, limiting real-time quality monitoring. Thus, constructing a soft sensor to accurately and in real-time predict MFR holds significant practical value. This study uses a polypropylene dataset to predict the reactor MFR (the primary variable) and includes seven auxiliary variables, with detailed descriptions and units in Table 2.

**Table 2 | Variables of the polypropylene production process**

| Variable | Description | Unit |
|---|---|---|
| V1 | Hydrogen Ratio | / |
| V2 | Reactor Pressure | bar |
| V3 | Reactor Bed Level | m |
| V4 | Liquefied Recycle gas to R-310 dome top | $Lh^{-1}$ |
| V5 | Hydrogen Flow | Kg/h |
| V6 | Reactor Temperature | K |
| V7 | Propylene flow | Kg/h |

Experimental settings

(1) Implementation details: All experiments were conducted on a server equipped with a 12th Gen Intel(R) Core(TM) i5-12600KF and 32GB RAM, utilizing Python for implementation. For the proposed method, unless otherwise specified, GPT-4o was employed as the LLM via API calls. To ensure consistency in output, the temperature was set to 0.

(2) LLM-ZAVS: For all datasets, we selected auxiliary variables comprising 50% of the total features to evaluate the performance of LLM-ZAVS. This was validated using linear regression and support vector regression. To ensure a fair comparison, we employed a fixed set of hyperparameters and 5-fold cross-validation for stable evaluation results. Due to inherent variability in LLMs outputs, we conducted five experiments to obtain five sets of LLM-ZAVS feature selection results, averaging them to derive the final outcome.

(3) The auxiliary variables selected by the LLM-ZAVS were used as input for LLM-UFSS. For LLM-UFSS-FSC (Methods), 200 samples were randomly chosen from the dataset to creat 10 contexts in SS-PT, each containing 20 text-based samples. From each context, 20 different test samples were randomly selected, ensuring no overlap through sampling without replacement. Each test sample underwent 10 experimental repetitions to establish prediction confidence intervals, with the average used as the final result. For LLM-UFSS-RAC (Methods), the 200 samples from LLM-UFSS-FSC were used to construct the IPDVS, matching each test sample with similar ones from the IPDVS to form the context. Due to its superior predictive performance and lower uncertainty, each test sample in LLM-UFSS-RAC was evaluated only once. For other benchmark methods, a grid search was performed to find the optimal parameters.

(4) Evaluation metrics: In this study, four evaluation metrics were employed to assess the effectiveness and accuracy of the proposed model: Mean Absolute Error (MAE), coefficient of determination ($R^2$), Root Mean Square Error (RMSE), and Mean Absolute Percentage Error (MAPE).

$$\mathrm{MAE} = \frac{1}{n}\sum_{t=1}^{n}\left|y_t - y_t^p\right| \tag{1}$$

$$R^2 = 1 - \frac{\sum_{t=1}^{n}(y_t - y_t^p)^2}{\sum_{t=1}^{n}(y_t - \bar{y})^2} \tag{2}$$

$$\mathrm{RMSE} = \sqrt{\frac{1}{n}\sum_{t=1}^{n}(y_t - y_t^p)^2} \tag{3}$$

$$\mathrm{SMAPE} = \frac{1}{n}\sum_{t=1}^{n}\frac{\left|y_t - y_t^p\right|}{(|y_t| + |y_t^p|)/2} \times 100\% \tag{4}$$

where $N$ represents the number of test samples, and $y_t$ and $y_t^p$ denote the ground truth and predicted values, respectively. The MAE, RMSE, and MAPE metrics evaluate the error between the ground truth and predicted values, with smaller values indicating better predictive performance. The $R^2$ metric reflects the proportion of variance in the target variable that is explained by the model, with values closer to 1 indicating superior model fit.

## Analysis of LLM-ZAVS feature selection results

We compared the performance of LLM-ZAVS with six common feature selection methods in two case studies, IndPensim and Polypropylene. These methods include Recursive Feature Elimination (RFE)[46], pearson correlation[47], filtering by Mutual Information (MI)[48], spearman correlation[49], fisher score[50], and random feature selection. As shown in Table 3, the text-based LLM-ZAVS achieved optimal or near-optimal results compared to these data-driven methods. Notably, LLM-ZAVS operates in a zero-shot manner, meaning it does not require direct access to data samples. This highlights the competitive advantage of LLM-ZAVS, which leverages real-world knowledge reasoning over numerical statistical analysis. Furthermore, LLM-ZAVS improved the linear regression MAE by 54.41% and 44.69% on IndPensim and Polypropylene, respectively, compared to random feature selection, demonstrates its effectiveness in enhancing model predictive performance and strong applicability.

**Table 3 | Comparative results of feature selection using various methods**

| Method | IndPensim | | | | Polypropylene | | | |
| --- | --- | --- | --- | --- | --- | --- | --- | --- |
| | LR | | SVR | | LR | | SVR | |
| | MAE | $R^2$ | MAE | $R^2$ | MAE | $R^2$ | MAE | $R^2$ |
| Random | 0.136 | 0.640 | 0.065 | 0.936 | 0.189 | 0.111 | 0.145 | 0.308 |
| Fisher | 0.089 | 0.859 | 0.062 | 0.927 | 0.151 | 0.368 | 0.110 | 0.546 |
| Spearman | 0.083 | 0.876 | 0.060 | 0.949 | 0.142 | 0.425 | 0.098 | 0.649 |
| Pearson | 0.064 | 0.915 | 0.057 | 0.950 | 0.143 | 0.420 | 0.105 | 0.601 |
| MI | 0.079 | 0.896 | 0.057 | 0.952 | 0.142 | 0.425 | **0.098** | **0.649** |
| RFE | **0.046** | **0.096** | 0.055 | 0.955 | 0.143 | 0.420 | 0.105 | 0.601 |
| LLM-ZAVS | 0.062 | 0.926 | **0.055** | **0.955** | **0.142** | **0.429** | 0.100 | 0.634 |

Even with identical prompts, using LLMs as feature selectors may result in different auxiliary variables being output. To assess the generative consistency of LLM-ZAVS, we propose a novel evaluation metric called the Average Selection Consistency Score (ASCS). Assuming LLM-ZAVS selects $m$ auxiliary variables from a set of candidates over $n$ repeated experiments, ASCS can be defined as follows:

$$\text{ASCS} = C_n^2 \times \sum_{i=1}^{n} \sum_{j=i+1}^{n} \left( \frac{|A_i \cap A_j|}{m} \right) \tag{5}$$

Where $C_n^2$ represents the binomial coefficient, and $|A_i \cap A_j|$ denotes the number of common elements between the auxiliary variable selection results of the $i$-th and $j$-th experiments. The ASCS value ranges from 0 to 1, with higher values indicating greater generative consistency of the feature selector. The ASCS results for LLM-ZAVS from five experiments are shown in Table 4. For IndPensim, where 11 auxiliary variables were selected, the ASCS is 0.75, meaning that, on average, 8.25 auxiliary variables were consistently selected across outputs. For Polypropylene, with four auxiliary variables selected, the ASCS is 1, indicating complete consistency across all five outputs. This demonstrates that LLM-ZAVS provides highly consistent feature selection results, exhibiting strong robustness.

**Table 4 | Average selection consistency score for LLM-ZAVS**

| Case Study | ASCS |
|---|---|
| IndPensim | 0.75 |
| Polypropylene | 1 |

After feature selection with LLM-ZAVS, we identified 11 variables from the IndPensim dataset: dissolved oxygen concentration, aeration rate, oxygen uptake rate, carbon evolution rate, sugar feed rate, temperature, pH, substrate concentration, acid flow rate, vessel volume, and generated heat. From the Polypropylene dataset, 4 variables were selected: hydrogen ratio, reactor pressure, hydrogen flow, and reactor temperature. These variables served as inputs for LLM-UFSS to implement soft sensing.

## Analysis of LLM-UFSS-FSC results

To validate the effectiveness of LLM-UFSS-FSC (Methods), four popular soft sensing algorithms were selected: Random Forest Regression (RFR)[51,52] and Multilayer Perceptron (MLP)[13], known for their robust nonlinear processing, k-Nearest Neighbors Regression (k-NN)[53] based on instance learning, and Principal Component Regression (PCR)[54,55], which combines PCA with multiple linear regression. Due to the extremely limited input training samples, deep learning networks were deemed unsuitable, so no comparison was made in this study. All comparative methods normalized and denormalized input data for optimal performance, whereas the LLM-UFSS-FSC utilized raw, unprocessed data.

The quantitative comparison results in Table 5 show that the LLM-UFSS-FSC achieved the lowest MAE and RMSE across both datasets without parameter updates or model modifications, outperforming traditional machine learning and neural network models. Specifically, on the IndPensim and Polypropylene datasets, the MAE for LLM-UFSS showed a reduction of 7.37% and 22.17%, respectively, compared to the second-best model, RFR. These findings also demonstrate that LLM-UFSS-FSC, leveraging text-based input, effectively learns from context samples and produces competitive results without normalization, exhibiting strong robustness to data scale variations.

**Table 5 | Performance comparision of LLM-UFSS-FSC with other methods**

| Methods | IndPensim | | Polypropylene | |
|---|---|---|---|---|
| | MAE | RMSE | MAE | RMSE |
| PCR | 6.227 | 7.290 | 0.547 | 0.776 |
| KNN | 4.898 | 6.197 | 0.560 | 0.615 |
| MLP | 4.205 | 5.683 | 0.524 | 0.622 |
| RFR | 3.513 | 4.944 | 0.415 | 0.522 |
| **LLM-UFSS-FSC** | **3.254** | **4.547** | **0.323** | **0.454** |

To visually demonstrate the model's fitting performance, we analyzed the first 80 test samples. Fig. 6 shows the actual and predicted curves for IndPensim using different methods. Each set of 20 test samples (separated by gray vertical dashed lines) uses the same 20 training samples for contextualization (LLM-UFSS-FSC) or training (other methods). The results indicate that the predictive curve of LLM-UFSS-FSC more accurately fits the true curve (Fig. 6e), with RFR performing second best (Fig. 6d), while PCR exhibits significant errors (Fig. 6a), suggesting its difficulty in handling complex nonlinear data. Notably, in the interval of test samples 60-80, other numerical methods show substantial errors and fluctuations, likely due to overfitting from the limited sample size. In contrast, LLM-UFSS, using the same 20 samples for SS-PT context, effectively leverages its analytical and reasoning capabilities to accurately track the true penicillin concentration values, thereby addressing model generalization issues and achieving precise predictions.

Traditional data-driven soft sensors produce fixed, deterministic predictions once trained, but they can yield high-confidence erroneous outputs when faced with uncertainties like extreme or noisy data[56]. Additionally, these models are vulnerable to adversarial attacks[57,58], impacting their reliability. They also lack the capability to quantify uncertainty, which is essential for industrial control decisions and safety. In contrast, LLMs are probabilistic generative models, generating each token through conditional probability distributions. By conducting multiple experiments and collecting diverse prediction results, LLMs can provide an uncertainty-aware prediction confidence interval.

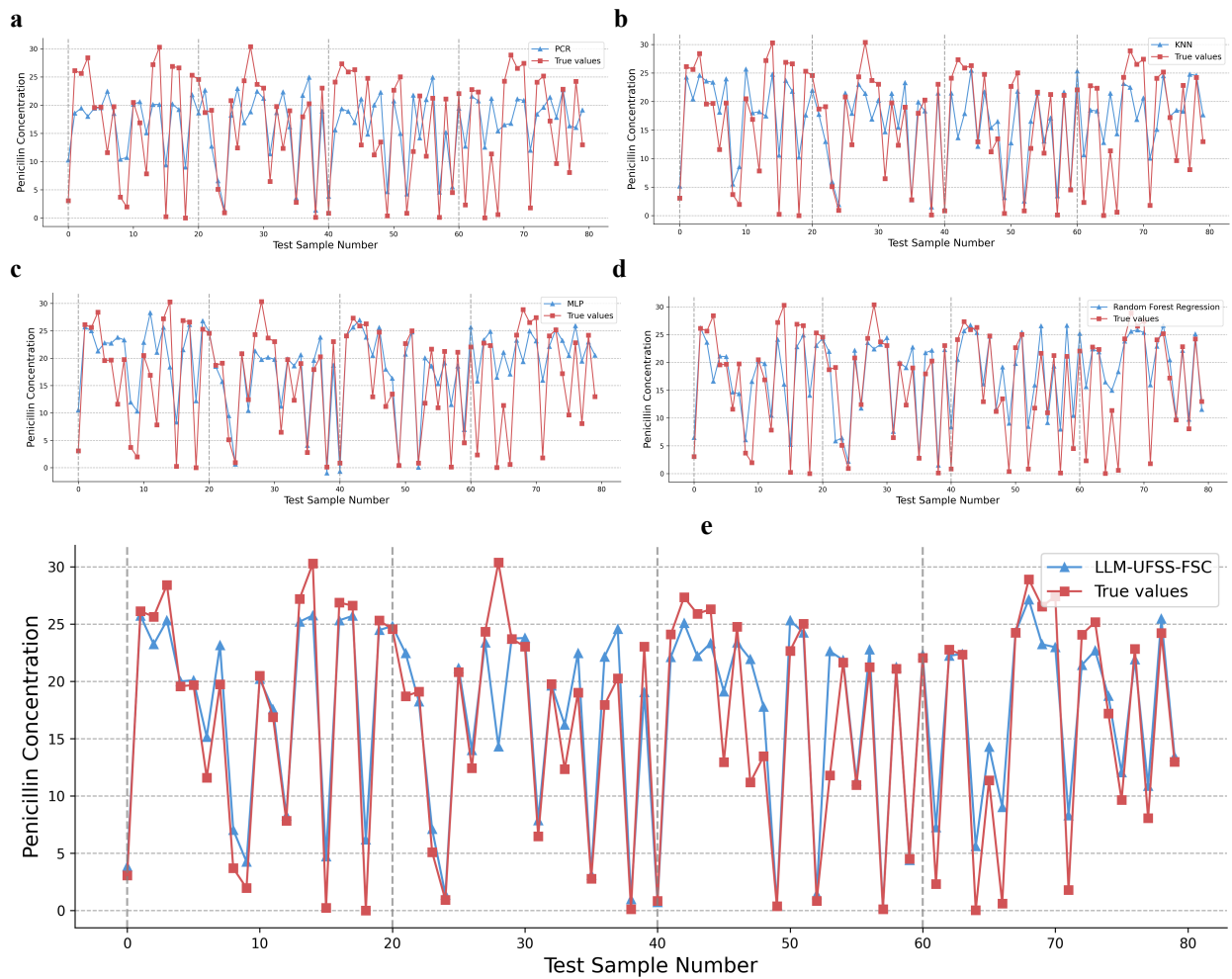

**Fig. 6 | Comparison of prediction results for different soft sensors on the test dataset.** Each subfigure illustrates the penicillin concentration predictions compared to true values across test samples. **a**, PCR. **b**, KNN. **c**, MLP. **d**, Random forest regression. **e,** LLM-UFSS-FSC.

We conducted 10 repeated experiments for each test sample in LLM-UFSS-FSC. Fig. 7 show uncertainty quantification visualizations for IndPensim (Fig. 7a) and Polypropylene (Fig. 7b). The light blue background represents the true values of the 20 context samples, while the light red background illustrates the LLM-UFSS-FSC's predicted fitting curve and its uncertainty confidence interval. The blue line indicates the true value curve, and the red line denotes the mean prediction from the experiments. Other colored dashed lines indicate the primary variable predictions from three random experiments, while the dark red area depicts the 95% confidence interval of LLM-UFSS-FSC predictions. The figures reveal that a smaller confidence interval corresponds to a better fit between the prediction curve and true values, while a larger interval suggests greater uncertainty and higher prediction error. By utilizing confidence intervals, not only can an approximate prediction range be provided, but the size of the interval can also be used to effectively quantify model uncertainty, offering comprehensive information to enhance decision-making quality.

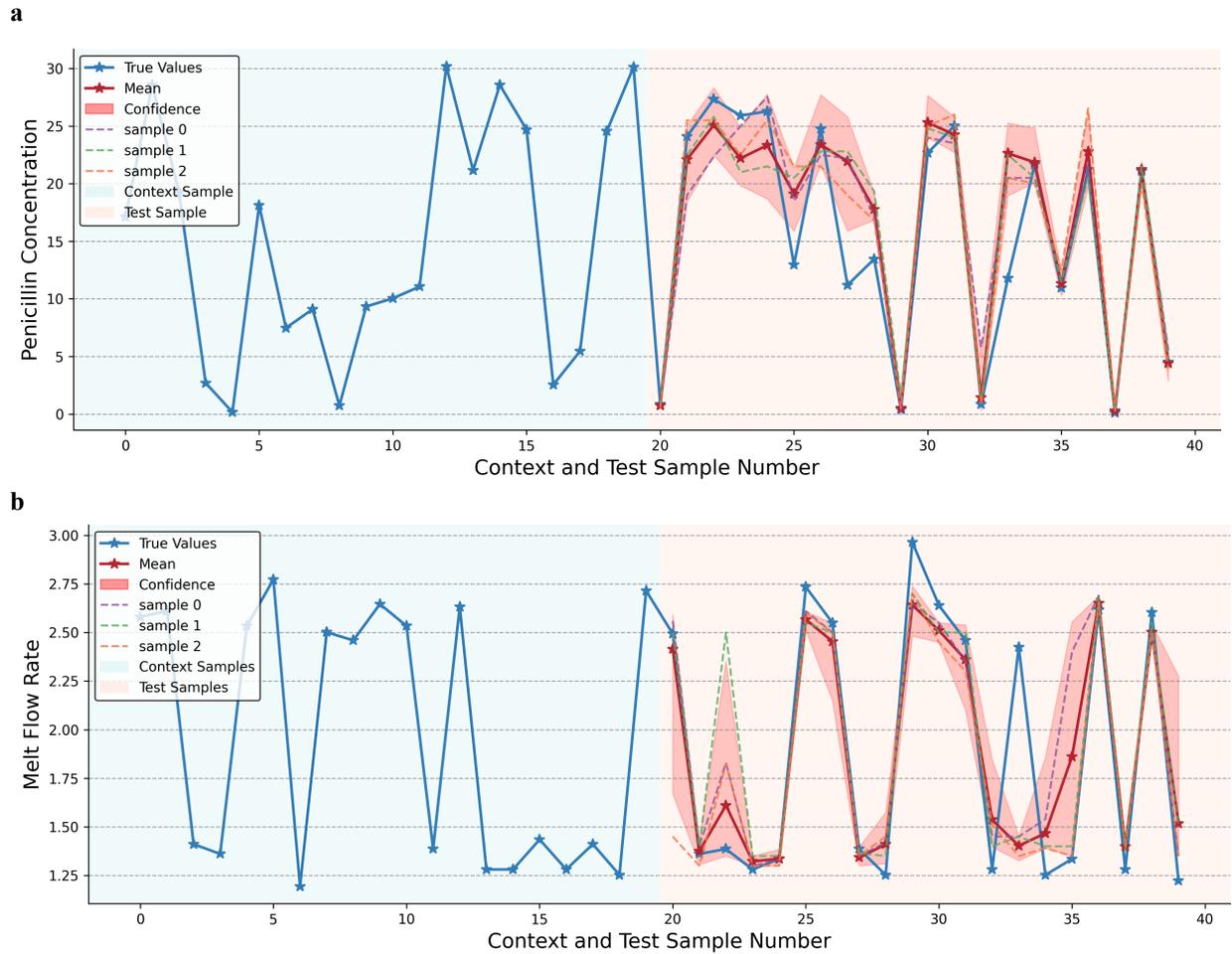

**Fig. 7 | Example of LLM-UFSS-FSC prediction uncertainty visualization results. a**, Penicillin concentration predictions with associated confidence intervals. **b**, Melt flow rate predictions with confidence intervals.

## Analysis of LLM-UFSS-RAC results

**Table 6 | Performace comparison of LLM-UFSS-RAC with other methods**

| Methods | IndPensim | | | | | | Polypropylene | | | | | |
|---|---|---|---|---|---|---|---|---|---|---|---|---|
| | MAE | | RMSE | | MAPE | | MAE | | RMSE | | MAPE | |
| N-Shot | 5 | 10 | 5 | 10 | 5 | 10 | 5 | 10 | 5 | 10 | 5 | 10 |
| PCR | 1.559 | 1.733 | 4.040 | 2.585 | 0.707 | 0.598 | 0.422 | 0.436 | 0.591 | 0.560 | 0.238 | 0.249 |
| KNN | 1.218 | 1.452 | 2.450 | 2.439 | 0.839 | 0.787 | 0.339 | 0.322 | 0.463 | 0.447 | 0.193 | 0.181 |
| MLP | 1.886 | 1.901 | 3.471 | 2.906 | 1.167 | 0.782 | 0.333 | 0.336 | 0.496 | 0.465 | 0.191 | 0.199 |
| RFR | 1.270 | 1.204 | 2.479 | 2.115 | 2.683 | 3.262 | 0.301 | 0.302 | **0.428** | **0.431** | **0.168** | 0.169 |
| LLM-UFSS-RAC | **0.905** | **0.651** | **2.535** | **1.200** | **0.451** | **0.360** | **0.300** | **0.253** | 0.494 | 0.438 | 0.177 | **0.148** |

In this section, we validate the performance of the LLM-UFSS-RAC (Methods), enhanced by IPDVS retrieval. We compare it with the four soft sensors in Fig. 6. To ensure a fair comparison, all context samples retrieved by LLM-UFSS-RAC are also used as training data for the other methods. However, using random training samples as model inputs results in poorer outputs for the comparative methods (see Table 5). The quantitative comparison results for all methods under 5-shot and 10-shot training samples are shown in Table 6. LLM-UFSS-RAC achieves competitive results, with a 10-shot MAE reduction of 45.93% and 16.23% compared to the second-best RFR across the two datasets, demonstrating the strong predictive capability of the proposed method.

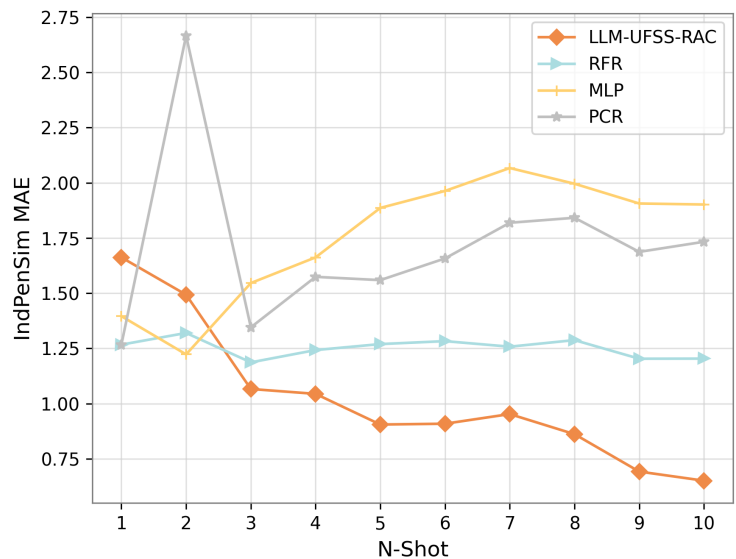

**Fig. 8 | The performance of different models with varying numbers of shots.**

Additionally, the MAE of LLM-UFSS-RAC with 10-shot context samples decreased by 28.07% and 15.67% compared to the 5-shot scenario, indicating that LLM can achieve more accurate predictions by considering more samples. To clearly illustrate the relationship between sample size and model performance, we visualized the MAE variations of different models under various N-shot conditions (Fig. 8). As the number of context samples increases, the prediction MAE of LLM-UFSS-RAC (orange line) consistently decreases, particularly during the initial 1-5 shot phase, where the reduction is most significant. This demonstrates that with more context samples, the proposed method effectively leverages ICL to integrate and understand multiple samples, enabling more accurate reasoning and showcasing its few-shot learning capability.

In the Fig. 8, the MAE of RFR stabilizes as the number of training samples increases, indicating that additional information from highly similar retrieved samples is limited and does not effectively enhance the model's predictive capability. The MAE for MLP and PCR even rises, suggesting potential overfitting or insufficient model complexity to handle subtle differences between similar samples. This further illustrates that LLM-UFSS-RAC possesses a finer-grained ability to perceive sample minor differences and make accurate decisions, partially addressing overfitting in numerical models.

To more intuitively reflect the prediction errors, we plotted the error box plots and absolute error curves (Fig. 9) for different models across all test samples. Fig. 9a and Fig. 9b show that the box plot for LLM-UFSS-RAC is significantly narrower, indicating a more concentrated error distribution and a smaller error range. Additionally, LLM-UFSS-RAC exhibits a lower median error. Fig. 10a and Fig. 10b further illustrate the visualization of absolute prediction errors for five soft sensors across all test samples. It is evident that the error curve for LLM-UFSS-RAC is generally lower and closer to zero. Moreover, the proposed method exhibits smaller error fluctuations, demonstrating superior stability and accuracy.

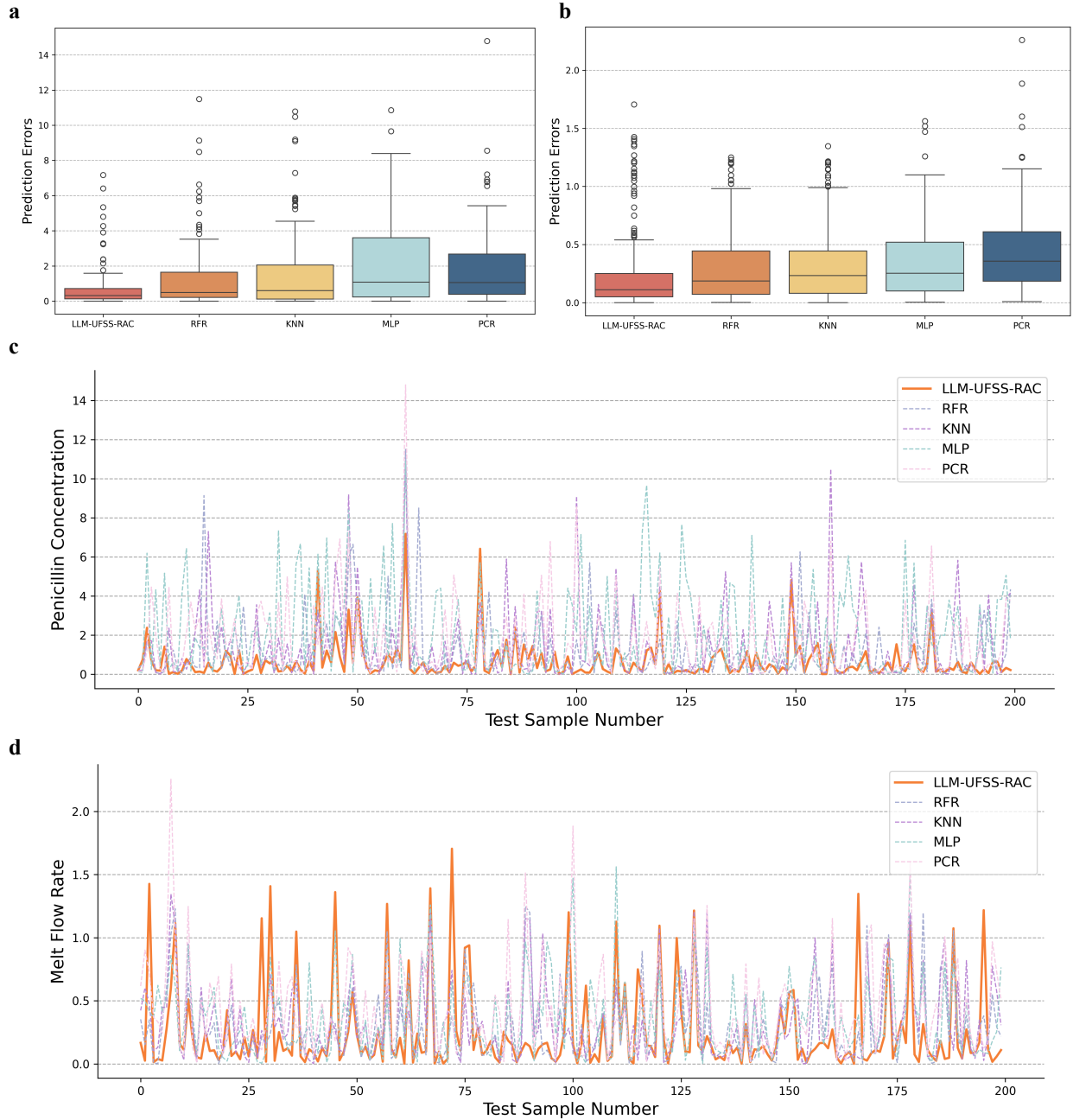

**Fig. 9 | Comparison of model prediction errors. a**, Box plots of errors for different methods on the IndPensim dataset. **b**, Box plots of errors for different methods on the Polypropylene dataset. **c**, Error curves of different methods across all test samples on the IndPensim dataset. **d**, Error curves of different methods across all test samples on the Polypropylene dataset.

## Analysis of robustness to missing values

In industrial settings, factors like data transmission anomalies, sensor failures, and environmental instability often lead to missing values in DCS-collected data[19,59]. However, data-driven soft sensors require strictly preprocessed, uniformly structured numerical data as input, making them ill-equipped to handle incomplete datasets[60]. Some soft sensors simply ignore samples with missing values, resulting in significant information loss, especially when the missing data rate is high. Others use imputation techniques to estimate these values, which increases the modeling complexity and development time[20].

**Table 7 | Performance comparision of different missing raitos**

| Missing Ratio (%) | IndPensim | | Polypropylene (10-shot) | |
|---|---|---|---|---|
| | MAE | RMSE | MAE | RMSE |
| 0% | 0.905 | 2.535 | 0.253 | 0.438 |
| 10% | 1.079 | 2.356 | 0.299 | 0.483 |
| 20% | 1.278 | 2.533 | 0.362 | 0.581 |
| 30% | 1.585 | 3.124 | 0.382 | 0.582 |
| 40% | 1.735 | 3.078 | 0.372 | 0.597 |
| 50% | 1.912 | 3.429 | 0.395 | 0.607 |
| **Average** | **1.416** | **2.843** | **0.344** | **0.548** |

The LLM-UFSS differs from traditional data-driven soft sensors by converting numerical inputs into prompt-based textual inputs, thus reducing the strict requirements on input format. To assess its applicability with missing data, we conducted five comparative experiments on two datasets using the LLM-UFSS-RAC 5-shot framework. We simulated missingness by masking auxiliary variables at rates from 10% to 50%, replacing missing data with 'N/A'. For instance, a sample like "Hydrogen Ratio: 0.17, Reactor Pressure: 30.576788, Hydrogen Flow: ..." becomes "Hydrogen Ratio: 0.17, Reactor Pressure: N/A, Hydrogen Flow: ...". Results in Table 7 show that while MAE and RMSE increase with higher missing rates, our method still achieves accurate predictions even with 50% missing data, with MAE as low as 1.912 and 0.344 for the two datasets, indicating robustness to missing data. Fig. 10 presents a histogram of prediction error distributions under varying missing rates, revealing that although the error distribution becomes more dispersed as the missing rate increases, it remains approximately unbiased and normally distributed, with most errors concentrated between -2 and 2, demonstrating strong predictive performance.

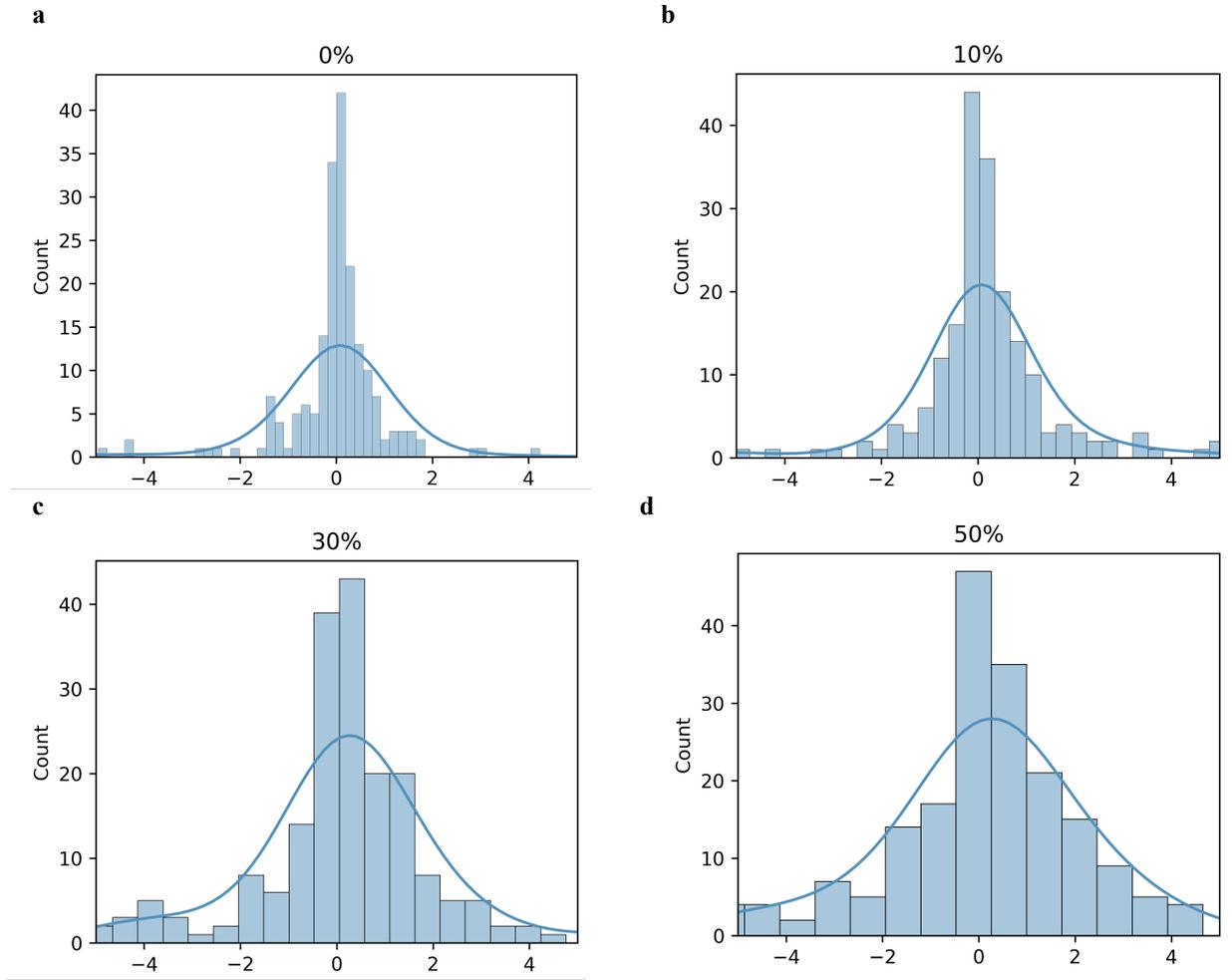

**Fig. 10 | Histogram of prediction errors for the IndPenSim dataset across different missing ratios. a**, 0% missing data. **b**, 10% missing data. **c**, 30% missing data. **d**, 50% missing data.

## Confidence score output analysis

In addition to constructing confidence intervals, we also explored an alternative approach for uncertainty perception in soft sensor predictions. This involves using a prompt strategy to instruct LLMs in SS-PT to perform confidence elicitation. By analyzing contextual information, the model outputs a confidence score ranging from 0 to 1, where scores closer to 1 indicate higher confidence. This approach aims to enhance risk assessment and error mitigation.

To validate the LLM-UFSS's ability to perceive uncertainty, we compared the average confidence scores across two datasets with data missing rates from 0% to 50%. As shown in Fig. 11a, confidence scores consistently decrease with higher missing rates, indicating that increased missing values reduce LLM-UFSS's confidence in prediction accuracy, thereby highlighting its capability to perceive prediction uncertainty. To further support this, we plotted the error and confidence score density for all samples in Fig. 11b. The horizontal axis represents the confidence scores, while the vertical axis represents prediction error, with values closer to 0 indicating better predictions. The figure reveals that the majority of confidence scores fall between 0.8 and 1, with fewer below 0.8. As confidence scores increase, the error visibly narrows towards the central value of 0. In the range of 0.9 to 1, the error is minimal, and predictive performance is optimal. This demonstrates that by guiding the model to output confidence scores, LLM-UFSS can effectively assess its own prediction accuracy.

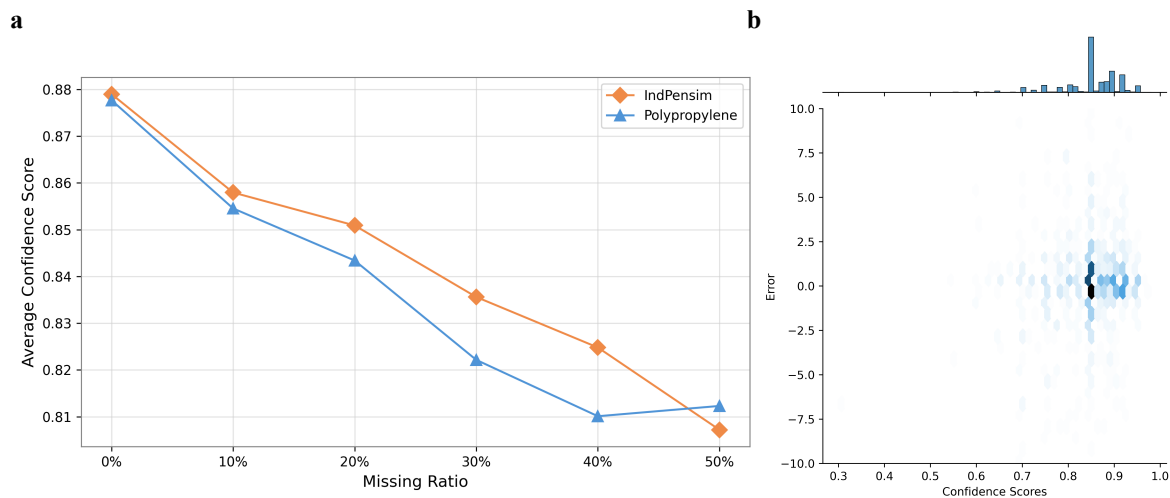

**Fig. 11 | Sensitivity analysis results of confidence scores. a**, Confidence scores of the IndPenSim dataset across various missing ratios. **b**, Density plot of confidence scores and error distribution.

## Analysis of explanation results

(1) Analysis of LLM-ZAVS explainability results

Through global-query, LLM-ZAVS generates importance scores and rankings for auxiliary variables, along with self-explanatory text regarding their feature importance. Fig. 12 illustrates the LLM-ZAVS 's outputs for global auxiliary variable identification and reasoning explanation across two datasets. The results, highlighted in orange, demonstrate that LLM-ZAVS effectively identifies and incorporates

a

> **[Explanation]:** To prioritize features for predicting Penicillin concentration, a thorough understanding of the process dynamics and the impact of various factors on penicillin biosynthesis is necessary. Here is a detailed explanation based on the given context:
>
> 1. **Dissolved Oxygen Concentration**: Oxygen is a critical factor in aerobic fermentation processes like penicillin production. Proper oxygenation ensures optimal microbial activity and penicillin yield. A low oxygen concentration can inhibit cell growth and reduce productivity. This makes it a crucial feature for predicting penicillin concentration.
> 2. **Substrate Concentration**: The availability of substrate directly influences the growth rate of Penicillium chrysogenum and its penicillin production. A higher substrate concentration can support more biomass, indirectly leading to higher penicillin production until metabolic limits or substrate inhibition occur.
> 3. **pH**: pH affects enzyme activity and overall microbial growth. The optimum pH must be maintained for the maximum production rate of penicillin; deviations can lead to reduced productivity.
> ...
> Based on the above considerations, each variable is scored based on its perceived importance to the overall process and production efficiency.
>
> **[Scoring and ranking]:** 1. Dissolved Oxygen Concentration - 1.00
> 2. Substrate Concentration - 0.95
> 3. pH - 0.93
> ...

b

> **[Explanation]:** To estimate the significance of each feature for predicting Melt Flow Rate (MFR) accurately, the following rationale was used, incorporating domain knowledge of the chemical processes involved in polypropylene production:
>
> 1. **Reactor Pressure**: Changes in the reactor pressure can directly impact the rate of polymerization and the molecular weight distribution of the polypropylene. It affects the solubility of gases in the polymer melt, influencing the MFR.
> 2. **Hydrogen Ratio**: Hydrogen acts as a chain transfer agent, significantly influencing the molecular weight and thus the MFR. Higher hydrogen ratios generally lead to lower molecular weight and higher MFR.
> 3. **Reactor Temperature**: The temperature in the reactor affects the kinetics of the polymerization reaction and the properties of the resulting polymer. Higher temperatures typically increase the MFR by accelerating the reaction rate.
> ...
> 7. **Liquefied Recycle gas to R-310 dome top**: This feature plays a role in overall reactor balance but has a less direct impact compared to the above features on MFR. Its influence is more on the efficiency and consistency of the process rather than on determining MFR directly.
>
> **[Scoring and ranking]:** 1. Reactor Pressure: 0.85
> 2. Hydrogen Ratio: 0.80
> 3. Reactor Temperature: 0.75
> ...

**Fig. 12 | Two examples of global feature selection explanations delivered by LLM-ZAVS. a,** Feature selection global explanation for the IndPensim dataset. **b,** Feature selection global explanation for the Polypropylene datase.

contextual embedded knowledge retrieved from the IKVS, providing professional interpretations of different variables based on this contextual information, as shown in blue.

Due to the challenge LLMs face in handling multiple tasks with fine granularity simultaneously, global-query may struggle to provide deeper reasoning for each variable. To address this, we introduce local-query for more detailed explanations, offering more comprehensive information. Fig. 13 presents two examples of local explanations generated by LLM-ZAVS.

a

> **[Local Explanation]:** The 'Sugar feed rate (Fs:L/h)' plays a pivotal role in the production of Penicillin due to several interconnected reasons rooted in the kinetics and biochemical pathways of the fermentation process.
> Here's a step-by-step analysis to understand its importance:
>
> 1. **Substrate Limitation and Growth**: Penicillium chrysogenum, the microorganism used for penicillin production, requires glucose as a primary carbon source. The concentration and rate at which sugar is fed (Fs) directly impact the growth rate of the biomass (mycelial concentration), which in turn affects the overall productivity.
> 2. **Fed-Batch Culture Strategy**: As documented in several studies, including those by Pirt (1974) and others, a fed-batch culture is often employed in industrial fermentations. This strategy controls the sugar feed rate to avoid substrate inhibition and maintain a quasi-steady-state, ensuring an optimal environment for penicillin synthesis.
> 3. **Metabolic Pathways and Kinetics**: The Michaelis-Menten kinetics and related equations (e.g., $-r_s = q_{s,max} C_s C_x G/(K_s + C_s)$) suggest that the substrate concentration (controlled by Fs) must be within a certain range to maximize specific sugar uptake rates without causing substrate inhibition.
> 4. **Simulation Insights**: The provided simulation data shows the impact of different sugar feed profiles on penicillin concentration. It is evident from the variance in penicillin production across different batches that sugar feed rate adjustments are crucial for maintaining optimal production levels.
> 5. **Optimization Studies**: Genetic Algorithms (GA) and other optimization techniques applied in the literature highlight that among various factors, substrate concentration and feed rate significantly improve penicillin yield. For example, the optimization results showed that setting the sugar feed rate to an optimal value (e.g., 1000 mol/hr) leads to substantial improvement in penicillin concentration.
> 6. **Empirical Models**: Empirical and kinetic models derived from multiple fermentation runs indicate that maintaining a stable and optimized sugar feed rate is essential for the consistent production of penicillin. Variations in Fs can lead to fluctuations in substrate concentration, which may adversely affect the production kinetics.
>
> In summary, the 'Sugar feed rate (Fs:L/h)' is crucial because it regulates the substrate availability, which directly influences the microbial growth, metabolic rates, and overall fermentation efficiency. Proper control of Fs ensures that the fermentation process operates within the optimal range, thereby maximizing penicillin production.

b

> **[Local Explanation]:** The oxygen uptake rate (OUR) is a crucial parameter in the fermentation process for penicillin production, particularly due to its direct association with the metabolic activity of P. chrysogenum. In the context provided, it is evident that the OUR is a significant determinant of the specific production rate of penicillin.
>
> Firstly, oxygen is essential for aerobic metabolism, and P. chrysogenum's production of penicillin is heavily reliant on aerobic respiration. The ability of the organism to efficiently convert nutrients into penicillin is closely tied to the availability of oxygen. The specific oxygen uptake rate correlates with the energy required for penicillin biosynthesis, as indicated by the requirement of approximately 1.6 mmol $O_2$/g-cells/hr for maximum productivity. This rate substantially exceeds the baseline rate at lower specific growth rates, reinforcing the need for ample oxygen to meet heightened metabolic demands during peak production.
>
> Moreover, optimal control of the oxygen supply can significantly impact the yield and productivity of the fermentation process. Studies have shown improved penicillin productivity with controlled oxygen feed rates, which align with the optimal specific uptake rates determined experimentally.
>
> The oxygen uptake rate also plays a role in maintaining cellular health and viability. Adequate oxygenation ensures the cells remain in the desired physiological state, which is critical for consistent penicillin production over prolonged fermentation periods. The relationship between specific oxygen uptake and penicillin productivity suggests that monitoring and adjusting OUR can lead to better control over the fermentation environment, thus optimizing penicillin yield.
>
> Therefore, the feature importance of the Oxygen Uptake Rate in predicting Penicillin concentration lies in its foundational role in energy metabolism, its direct impact on penicillin biosynthesis, and its utility in process optimization and control.

**Fig. 13 | Two examples of local variable explanations delivered by LLM-ZAVS. a**, Local explanation of the importance of sugar feed rate for penicillin concentration in the IndPensim dataset. **b**, Local explanation of the importance of oxygen uptake rate for melt flow rate in the Polypropylene dataset.

In Fig. 13a, the case illustrates a local explanation of the impact of the sugar feed rate on penicillin concentration in the IndPenSim dataset. Using the CoT technique in AVS-PT, LLM-ZAVS decomposes the problem for step-by-step analysis (as shown in blue). Compared to global explanations, local explanations provide more detailed insights into individual auxiliary variables. Notably, the orange text highlights how the introduction of RAG allows LLM-UFSS to effectively utilize and comprehend professional knowledge from the IKVS, including kinetic equations (points 2 and 6), simulation data (point 3), optimization experiment results (point 4), and empirical models (point 6). This approach enables more professional analysis and reasoning using external database knowledge, mitigating the hallucination phenomenon and enhancing the reliability of the explanations. Fig. 13b further supports this view with a local explanation of the oxygen uptake rate's effect on penicillin concentration.

(2) Analysis of LLM-UFSS explainability results

A major challenge for data-driven soft sensors during practical deployment is their lack of reliability. This stems from the difficulty in interpreting end-to-end trained black-box models, which hinders users from effectively assessing the reasonableness and trustworthiness of the outputs. To address this issue, we utilize the advanced emergent capabilities of LLMs, including natural language understanding, generation, and step-by-step reasoning, to provide clear, human-readable explanations for complex soft sensor prediction. We first analyze the explainability results of LLM-UFSS-FSC predictions. Fig. 14 presents textual explanations for soft sensor outputs across two datasets. The orange text indicates that the proposed method considers the auxiliary variable importance conclusions generated by LLM-ZAVS and the context samples. The purple text signifies the influence of CoT prompting technique. We divide the main explanation into a reasoning section (blue text) and a conclusion section (green text). The reasoning section demonstrates how LLM-UFSS uses ICL to position the auxiliary variable values within an approximate range, and the conclusion section then illustrates the specific impact of this range on the main variable's value, primarily derived from ICL or biochemical knowledge. Finally, the method integrates these factors to derive accurate predictions (e.g., the true penicillin value is 26.123 in Fig. 14a, and the true MFR value is 1.335 in Fig. 14b).

a

> **[Predicted Value]:** 26.309
> **[Confidence Score]:** 0.9
> **[Explanation]:** To predict the Penicillin concentration, I closely analyzed the feature importance scores and compared the provided sample with similar context examples. The sample has notable values for key features such as Dissolved Oxygen Concentration (9.638), pH (6.5181), Substrate Concentration (0.0012666), and Generated Heat (498.58). These variables are crucial for determining the Penicillin yield.
> 1. The Dissolved Oxygen Concentration of 9.638 is comparable to earlier samples with concentrations around 9.6-12.3, which typically showed intermediate to high Penicillin outputs ranging roughly between 20-29.
> 2. A pH value close to the average optimal range around 6.5 suggests effective enzymatic activity, which aids in maximizing Penicillin production.
> 3. The Substrate Concentration of 0.0012666 is within the common range seen in samples yielding high Penicillin concentrations (above 25).
> 4. The Generated Heat value of 498.58 is aligned with conditions where metabolic activity was high, contributing to productivity.
> Given these factors and referencing the example data, it is reasonable to predict the Penicillin concentration to be approximately 26.309, with a high confidence score of 0.9.

b

> **[Predicted Value]:** 1.35
> **[Confidence Score]:** 0.82
> **[Explanation]:** Given the importance of the auxiliary variables in predicting Melt Flow Rate (MFR), I used the following step-by-step approach to derive the prediction:
> 1. **Hydrogen Ratio (0.143):** This value is relatively low compared to samples with MFRs in the higher range (e.g., closer to 0.20 or higher). Lower Hydrogen Ratios tend to result in lower MFRs.
> 2. **Reactor Pressure (30.821402):** This pressure is somewhat consistent with the average reactor pressure in the dataset, but it is not exceedingly high or low. Thus, it's expected to have a neutral to slightly positive impact on MFR.
> 3. **Hydrogen Flow (0.166888492):** This is on the higher end of the values in the dataset. As such, it would typically contribute to maintaining chain transfer reactions, keeping the MFR in a moderate range.
> 4. **Reactor Temperature (78.534431):** The temperature is slightly lower than the common range (~79-80°C). Lower temperatures generally decrease the reaction rate but also contribute to higher molecular weights, leading to lower MFRs.
> After considering all these factors and comparing closely matching samples (Hydrogen Ratios around 0.14, Reactor Pressures around 30.8, Hydrogen Flows around 0.16, and Temperatures around 78-79°C), an MFR value close to 1.35 seems reasonable. The confidence score of 0.82 indicates a high degree of certainty but accounts for any minor variations.

**Fig. 14 | Two examples of soft sensor explanations delivered by LLM-UFSS-FSC. a**, Explanation results for penicillin concentration prediction in the IndPensim dataset. **b**, Explanation results for melt flow rate prediction in the Polypropylene dataset.

To validate the effectiveness of the IPDVS and the context-based retrieval augmentation methods proposed in LLM-UFSS-RAC (Methods), we present two explanation examples in Fig. 15. The ground truth values for the two cases are 19.67 (Fig. 15a) and 1.361 (Fig. 15b), respectively. The explanations in Fig. 15 further confirm the conclusions mentioned earlier. Unlike LLM-UFSS-FSC, LLM-UFSS-RAC enhances prediction by retrieving similar samples, positioning predicted values within a more precise range (e.g., points 2 and 3 in Fig. 15a). By analyzing relevant samples (indicated by the blue text in Fig. 15b), LLM-UFSS-RAC achieves finer-grained analysis, enhancing predictive performance.

**a**

> **[Predicted Value]:** 19.50
> **[Confidence Score]:** 0.85
> **[Explanation]:** Based on the context examples and the feature importance rankings:
> 1. **Dissolved Oxygen Concentration**: The given value (11.328) is close to the values in the context examples where the penicillin concentration ranged from 17.322 to 20.743. Since this feature has the highest importance score, it heavily influences the prediction. The closest example value to 11.328 is 11.309 which had a resulting penicillin concentration of 19.415.
> 2. **pH**: The given pH (6.508) is within the range of the context examples (6.4892 to 6.5049). pH is the second most important feature impacting penicillin concentration. The closest example value is 6.5049 with a penicillin concentration of 17.505. However, the slight increase in pH could push the concentration towards the higher end.
> 3. **Substrate Concentration**: The given value (0.0012116) is slightly higher than the average of the context examples (0.0011723 to 0.0012861). This feature also plays a significant role, given its importance score, leading to slight variations in penicillin concentration.
> 4. **Temperature, Aeration Rate, and Sugar Feed Rate**: These features are consistent across all examples, so they maintain stability in the predictions, implying their direct influence is constant at the given levels.
> Considering these key features and their values in the examples, a predicted penicillin concentration around 19.5 is reasonable. Given the slight variances but overall proximity to the historical data, a confidence score of 0.85 is assigned.

**b**

> **[Predicted Value]:** 1.335
> **[Confidence Score]:** 0.9
> **[Explanation]:** Examining the context examples provided and the importance scores for each feature, the sample to be predicted has features very similar to the first context example.
> The Hydrogen Ratio (0.193), Reactor Pressure (30.479473), Hydrogen Flow (0.183422727), and Reactor Temperature (80.074524) are closely matched with the first context example where the MFR was 1.335001067. Given that the Hydrogen Ratio is the most significant feature and it is exactly the same in the predicted sample, along with very minor variations in other features, the predicted MFR would closely match the first context example.
> Hence, the predicted Melt Flow Rate for the sample is 1.335 with a high confidence score of 0.9

**Fig. 15 | Two examples of soft sensor explanations delivered by LLM-UFSS-RAC. a**, Explanation results for penicillin concentration prediction in the IndPensim dataset. **b**, Explanation results for melt flow rate prediction in the Polypropylene dataset.

## Ablation study

To validate the effectiveness and contributions of each key component in the proposed method, we conducted ablation experiments with various variants of LLM-UFSS. The LLM is a critical component, so we compared the default GPT-4o[26] model with three other state-of-the-art LLMs (GPT-3.5-turbo[61], GPT-4[26], Gemini-1.5-pro-exp-0801[27]) across two datasets. The MAE results are shown in Fig. 16a. It is evident that GPT-3.5-turbo exhibits a significantly higher MAE of 2.132 on the IndPenSim dataset, indicating a substantial performance gap compared to other models. In contrast, Gemini-1.5-pro-exp-0801 achieves the lowest MAE of 0.798, a 62.57% reduction compared to GPT-3.5-turbo, highlighting the impact of LLMs selection on LLM-UFSS's predictive performance. Furthermore, we believe that Gemini-1.5-pro-exp-0801 offers superior mathematical analysis capabilities compared to the GPT-4 series.

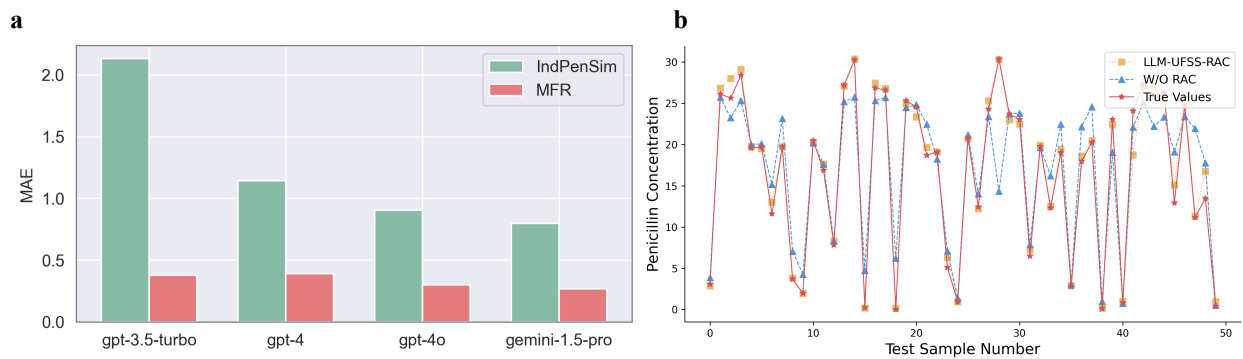

**Fig. 16 Ablation study analysis results. a**, Comparison of pediction prformance with dfferent LLMs. **b**, Comparative analysis of predictive performance between LLM-UFSS-RAC and W/O RAC.

To validate the role of IPDV and vector retrieval enhancement in ICL, we conducted an ablation study by removing these components from LLM-UFSS-RAC, denoted as W/O RAC. The fitting performance of both models on the test set is illustrated in Fig. 16b. The results indicate that LLM-UFSS-RAC demonstrates superior fitting performance, suggesting that ICL can achieve more accurate predictions through the use of enhanced similar samples.

**Table 8 | Comparison of confidence scores**

| Methods | Average Confidence Score | |
|---|---|---|
| | IndPensim | Polypropylene |
| W/O RAC | 0.860 | 0.865 |
| **LLM-UFSS-RAC** | **0.882** | **0.878** |

To evaluate the effectiveness and sensitivity of the confidence scores output by LLM-UFSS, we compared the average confidence scores of W/O RAC and LLM-UFSS-RAC across two datasets (Table 8). W/O RAC uses random samples for context, while LLM-UFSS-RAC employs enhanced similar samples. The results show that LLM-UFSS-RAC achieves higher average confidence scores, indicating that the confidence score effectively perceives the context samples and assesses the reliability of its own predictions based on the degree of perception. The confidence score aligns with factual patterns and holds significant reference value.

**Table 9 | Results of the ablation study on prompts**

| Model | LLM-UFSS-FSC | | LLM-UFSS-RAC | |
|---|---|---|---|---|
| | MAE | RMSE | MAE | RMSE |
| W/O EC | 4.410 | 10.990 | 1.117 | 2.694 |
| W/O Role | 3.694 | 5.518 | 1.005 | 2.663 |
| W/O CoT | 3.731 | 5.663 | 0.953 | 2.771 |
| With All | **3.254** | **4.547** | **0.905** | **2.535** |

To evaluate the effects of various prompt components, we conducted ablation experiments using three prompt variants on both LLM-UFSS-FSC and LLM-UFSS-RAC: (1) W/O Role: removing the role component from SS-PT, (2) W/O CoT: excluding chain-of-thought instructions, and (3) W/O EC: removing instructions related to output explanations and confidence scoring. The results, shown in Table 9, indicate that removing any component leads to a decline in the overall performance. The first variant demonstrate that specifying the LLM's role within the prompt enables more accurate predictions through role-playing. The second variant, along with results from Fig. 14, shows that the CoT technique incrementally guides the LLM in problem-solving, thereby enhancing task performance. In the third variant, where the LLM is no longer required to explain and score its predictions, a noticeable drop in predictive performance is observed. This suggests that self-explanation and uncertainty assessment via instructions encourage deeper reasoning in LLMs, ultimately improving predictive capabilities.

## Discussion

This paper introduces LLM-FUESS, the first two-stage soft sensing framework based on the ICL paradigm of LLMs, designed to overcome various challenges in traditional data-driven soft sensors. In the first stage, the LLM-ZAVS retrieves domain-specific knowledge from the IKVS and integrates it with the internal knowledge of LLMs. This integration endows LLMs with expert-level analytical capabilities for auxiliary variable selection. In the second stage, LLM-UFSS employs minimal samples as context task demonstrations, leveraging the powerful ICL capabilities to achieve robust predictive performance without any model training or parameter updates. Additionally, by constructing the IPDVS and introducing the RAG, we offer an alternative strategy for enhancing context samples when ample data is available. We exploit the robust text generation and probabilistic features of LLMs in both stages to provide human-readable, explainable insights, and for LLM-UFSS, we design two methods to quantify prediction uncertainty as a basis for evaluation. To enhance the task adaptability and usability of LLM-FUESS across diverse scenarios, we introduce various prompting strategies, creating two highly encapsulated fill-in-the-blank templates: AVS-PT and SS-PT for each stage, respectively.

Extensive experiments were conducted on datasets from two distinct domains, IndPensim and Polypropylene. The results from the first stage demonstrate that LLM-ZAVS can consistently and effectively select auxiliary variables, providing explanations from both local and global perspectives with a high degree of professionalism. In the second stage, the experiments reveal that LLM-UFSS not only achieves competitive performance compared to data-driven soft sensors but also exhibits greater robustness and flexibility regarding input data formats, quantities, and types. Furthermore, the experiments indicate that LLM-UFSS possesses strong self-explanation and uncertainty awareness capabilities, enhancing the method's transparency and risk awareness. This provides practitioners with more comprehensive and valuable decision-making information.

## Methods

### Overall design

As the parameters and training corpus of large LLMs continue to expand, they exhibit emergent capabilities—abilities not present in smaller-scale models. This paper proposes a novel few-shot soft sensor method, LLM-FUESS (Uncertainty-Awareness and Self-Explanation), leveraging these emergent abilities of LLMs. Typically, data selection and soft sensor modeling are considered two independent components of the soft sensor pipeline. As illustrated in Fig. 1, the proposed LLM-FUESS consists of two stages: (a) LLM-ZAVS and (b) LLM-UFSS, which perform feature selection and soft sensing tasks, respectively. These stages are detailed in "LLM-based zero-shot auxiliary variable selector (LLM-ZAVS)" and "LLM-based uncertainty-aware few-shot soft sensor (LLM-UFSS)". The meticulous design of prompt strategies is crucial for eliciting the emergent capabilities. Therefore, we have developed two fill-in-the-blank prompt templates, AVS-PT and SS-PT (AVS-PT and SS-PT with prompt engineering), for these stages to maximize the reasoning abilities and performance of the models.

## LLM-Based zero-shot auxiliary variable selector (LLM-ZAVS)

In the context of expanding production scales and increasingly complex industrial processes, the dimensionality of process variables has significantly increased. Redundant variables can impair the predictive performance and computational efficiency of soft sensor. By selecting an appropriate subset of auxiliary variables, we can eliminate redundancy and conserve LLMs input tokens. Additionally, identifying key quality variables supports product quality control and enhances analysis, interpretation, and prediction by LLMs. To address this, we propose a novel Zero-Shot Auxiliary Variable Selector (LLM-ZAVS). As depicted in Fig. 2, LLM-ZAVS comprises three main components: (a) construction of the industrial knowledge vector store, (b) filling in the prompt template, and (c) generation of explainable auxiliary variable selection results.

In component (a), we construct a dynamic external industry knowledge vector store tailored to the current task using RAG. This approach address the domain-specific knowledge gaps of LLMs in knowledge-intensive industrial tasks, enhancing variable selection capabilities and reducing hallucinations. Initially, we collect relevant documents from various data sources, including books, research papers, web pages, and technical reports, to compile an unified, authoritative internal industrial scenario knowledge base, denoted as $D_{kb}$. The text within $D_{kb}$ is then split into smaller document chunks:

$$D_{kb} = \{D_1^{kb}, D_2^{kb}, ..., D_n^{kb}\} \tag{6}$$

Subsequently, we employ an embedding model $f_e$ to encode each chunk $D_i$ into high-dimensional vectors:

$$f_e : D_i^{kb} \rightarrow V_i \in \mathbb{R}^d \tag{7}$$

where $d$ represents the embedding dimension. The resulting embedding vectors $V = \{V_1, V_2, ..., V_n\}$ are stored in the Industry Knowledge Vector Store (IKVS) to facilitate efficient retrieval.

In step (b), we populate the auxiliary variable selection prompt template (AVS-PT) to creat an auxiliary selection query for the LLMs. The AVS-PT (Fig. 2a) consists of three components: elements (in blue), a fixed template (in black), and task-specific configurations (in green). First, the user inputs task-specific information—such as the industrial process, facility, number of variables, and primary variables—into the AVS-PT, replacing the placeholders to form a basic prompt query (see "AVS-PT and SS-PT with Prompt Engineering" for more details). Next, we retrieve relevant knowledge from an external database to integrate into the AVS-PT as context, enhancing the LLM's generation process. To accommodate different generation requirements, we have designed two retrieval-enhanced generation paths: the global query (blue dashed line) and the local query (yellow dashed line). In the global query, the user inputs all candidate auxiliary variables names into the IKVS as a query, embedding them into high-dimensional query vectors $E_q \in \mathbb{R}^d$ using the embedding function $f_e$. We then calculate the euclidean distance between $E_q$ and all vectors in the IKVS to determine vector similarity:

$$d(E_q, V_i) = \|E_q - V_i\|_2 = \sqrt{\sum_{j=1}^{d}(E_q^j - V_i^j)^2} \qquad (8)$$

Finally, the top $K$ most relevant vectors are identified, and the corresponding documents $T = \{T_1, T_2, ..., T_K\}$ are retrieved as context for the AVS-PT, resulting in the final auxiliary variable selection query (global). The AVS-PT (global) instructs LLMs to generate scores and importance rankings for all auxiliary variables, along with detailed explanations. In step (c), we input the global query into the pre-trained LLM $M_{frozen}$, with all parameters frozen. Using LangChain, we format the LLM's output into JSON, obtaining a stable importance ranking $r_j$ and scoring $s_j$, as well as a global explanation (blue box) $exp_g$:

$$s_j, r_j, exp_g = M_{frozen}(AVS\text{-}PT(t, c_g)), j \in \{1,...,l\} \qquad (9)$$

where $t$ represents task-specific configurations, and $c_g$ denotes the relevant documents retrieved by the global query. In the local query, users select a specific candidate auxiliary variable as the query for the IKVS, and through the same retrieval-augmented process as the global query, generate detailed feature importance explanations $exp_l$ for the candidate auxiliary variable (orange box):

$$exp_l = M_{frozen}\left(AVS\text{-}PT(t, c_l)\right) \tag{10}$$

Where $c_l$ represents the relevant documents retrieved by the local query. Without any model parameter adjustments or prior data inspection, the external knowledge augmentation from both global and local queries, along with the LLM's robust reasoning capabilities, enables the generation of importance rankings for auxiliary variables and provides detailed text-based self-explanations of the decision-making process, offering high readability and reference value.

## LLM-based uncertainty-aware few-shot soft sensor (LLM-UFSS)

After processing by the first-stage LLM-ZAVS module, we filter a set of auxiliary variables most crucial to the primary variables. In the second stage, these variables serve as inputs for predicting primary variable values through a soft sensor. Unlike previous approaches, our LLM-Based Uncertainty-Aware Few-Shot Soft Sensor (LLM-UFSS) requires no model training or gradient updates. Instead, it utilizes ICL with prompt engineering, using a few examples to leverage the LLM's capabilities to analyze data and prediction. Furthermore, compared to earlier soft sensors that could only generate singular numerical predictions, the LLM-UFSS also provides detailed reasoning explanations, confidence scores, and confidence intervals for uncertainty quantification of the generated results.

The LLM-UFSS pipeline, similar to LLM-ZAVS, consists three parts: (a) Construction of the industrial process data vector store, (b) Filling in the prompt template, and (c) Generation of explainable soft sensor results (Fig. 3). With the widespread application of DCS, collecting and storing large amounts of process variable data has become feasubke. Inspired by RAG, the stage (a) aims to construct a retrievable industrial process data vector store to provide LLMs with more valuable contextual examples for improved predictions. Initially, we obtain raw historical process variable data through DCS applications, available in formats such as JSON, Excel, or CSV. We then filter this data using the LLM-ZAVS, selecting the auxiliary variables and removing less important, redundant, and spurious variables to form the DCS process variable database $D_{pv}$. Subsequently, we segment the database according to sample time steps, resulting in $n$ independent data samples:

$$D_{pv} = \left\{ D_1^{pv}, D_2^{pv}, ..., D_n^{pv} \right\} \tag{11}$$

where $D_i^{pv} = \left( x_i^1, x_i^2, ..., x_i^m, y_i \right)$, $m$ is the number of auxiliary variables, $x_i$ denotes the auxiliary variable values, and $y_i$ is the true value. Following the LLM-ZAVS, each sample $D_i^{pv}$ is encoded into high-dimensional vectors for indexing and storage in the industrial process data vector store (IPDVS).

In phase (b), we construct a Soft Sensor Prompt Template (SS-PT) tailored for the soft sensing task (Fig. 4). Users input task-specific industrial background information $t$ to configure the SS-PT. Additionally, feature importance rankings $r_j$ and scores $s_j$ generated by LLM-ZAVS, along with global feature explanations $\exp_g$, are incorporated as extra contextual information. This provides the LLMs with finer-grained prompts, enabling more in-depth and complex reasoning. It is essential to provide high-quality sample examples for ICL inference and generalization, so we designed two pathways to supply ICL samples based on varying industrial scenarios. When on-site samples are minimal and insufficient to construct the IPDVS, LLM-UFSS employs the available few-shot samples as inputs into the SS-PT, termed LLM-UFSS Few-Shot Contextualization (LLM-UFSS-FSC). Conversely, if the DCS gathers substantial amount of data, a text-based test sample—comprising auxiliary variable names and values—queries the IPDVS to retrieves several samples with high similarity to the test sample, which are then input into the SS-PT (orange dashed line), known as LLM-UFSS Retrieval-Augmented Contextualization (LLM-UFSS-RAC). These context samples are denoted as $I$, with a total of $k$ examples, where $I$ is defined as follows:

$$I = f\left(x_1^1, x_1^2, \ldots, x_1^m, y_1\right), f\left(x_2^1, x_2^2, \ldots, x_2^m, y_2\right), \ldots, f\left(x_k^1, x_k^2, \ldots, x_k^m, y_k\right) \quad (12)$$

where $f(\cdot)$ represents a text formatting function that converts numerical samples into a textual format consistent with the test samples (as indicated by the purple dashed line). Upon constructing the SS-PT, a complete input prompt for the LLMs, referred to as the soft sensor query, is obtained. In phase (c), this query is input into the pre-trained LLMs, which analyze the context samples through ICL to output the soft sensor predictions $y_p$ for the primary variables of the test samples.

We have incorporated self-explanation instructions into the SS-PT, leveraging the LLM's robust reasoning and natural language generation capabilities to produce clear, detailed, and human-readable explanations of model decisions ($exp_{ss}$). This aids operators in making safer, more informed decisions and conducting causal analysis. Beyond precise point estimates, we have developed two distinct methods for quantifying output uncertainty. Firstly, by selecting the next token from multiple high-probability options, the LLM generates diverse textual outputs, allowing us to construct a predictive confidence interval through repeated experiments. Analyzing the width and boundaries of this interval helps users assess the model's reliability and stability, facilitating more informed decision-making. Secondly, we introduce instructions in the SS-PT to compel the LLMs to generate a confidence score for its predictions, reflecting the model's confidence level and enabling awareness of prediction performance uncertainty. Lastly, our model also exhibits strong robustness to input data uncertainty. Specifically, since the LLMs formats numerical inputs as text-based prompts, any missing values in data samples can be replaced with 'N/A'. Compared to traditional numerical models that require strict normalization and imputation of missing values, LLM-UFSS simplifies the data preprocessing workflow and avoids inappropriate handling that could lead to information loss or distortion. This approach enhances readability, robustness, flexibility, and intelligence. Analysis of ablation experiments shows that prompting the LLMs for self-explanations and confidence scores also improves the predictive performance of the proposed method.

## AVS-PT and SS-PT with prompt engineering

By incorporating key elements through prompt engineering techniques, we develop a fixed fill-in-the-blank template that allows users to simply input external information into the designated placeholders, generating a complete task-specific prompt. The structure and content of the prompt template significantly influence the model's output. For the distinct requirements of the LLM-ZAVS and LLM-UFSS modules, we have devised two carefully crafted prompt strategies: the Auxiliary Variable Selection Prompt Template (AVS-PT) and the Soft Sensor Prompt Template (SS-PT) (Fig. 4). These templates are designed to activate and guide the LLM's reasoning process effectively.

In designing the AVS-PT, we identified five key components: role, data, instruction, context, and main user prompt. 1) Role: This element assigns a specific role to the LLMs, allowing it to dynamically adapt to the task and context for more accurate responses. In our study, the LLMs acted as experienced industrial data analysts. Considering the complexity and potential risks in industrial environments, where decision-making greatly impacts safety, we employed emotional stimuli[44] to emphasize the impact of analysis on industrial safety. 2) Data: This provides detailed background information on the data collection process, enabling LLMs to gain a comprehensive understanding of the data context. Users are required to input key information into the fill-in-the-blank template, including the specific industrial process, equipment, primary variable, and the number of features. 3) Instruction: We have designed two distinct sets of instructions for local and global queries. Local queries focus on the importance of a specific auxiliary variable, while global queries rank all auxiliary variables. The CoT prompting technique enhances the accuracy and transparency of LLMs in industrial decision-making by guiding step-by-step reasoning, resulting in detailed self-explanatory texts on feature importance scores. 4) Context: LLMs have limited ability to handle knowledge-intensive industrial tasks due to a lack of domain-specific knowledge. To address this, we have constructed a IKVS from external sources to expand the foundational knowledge of LLMs. The relevant knowledge retrieved from IKVS is incorporated into this element to enhance their reasoning and self-explanatory capabilities. 5）Main User Prompt: This element specify the variable names to be analyzed and instruct LLMs to generate responses. For local queries, the auxiliary variable name under analysis replaces the placeholder {Auxiliary Variable}, while for global queries, all auxiliary variable names are inputted, replacing {Auxiliary Variables}.

For the development of a few-shot soft sensor, we designed the SS-PT template, which consists of seven components: role, data, feature importance score and ranking, global explanation, instruction, context prompt, and main user prompt. 1) Role: identical to that in AVS-PT. 2) Data: Similar to AVS-PT, but includes all auxiliary variables identified by LLM-ZAVS. 3) Feature Importance Score and Ranking: Generated by LLM-ZAVS, enabling LLMs to perform more granular soft sensor reasoning. 4) Global Explanation: Also generated by LLM-ZAVS. 5) Instruction: Guides LLMs to execute ICL to generate predictions and produce a reasonable explanation based on the CoT. 6) Context Prompt: Formats few-shot labeled input-output pairs into a text sequence and inserted into this element. 7) Main User Prompt: Users input the test samples into this element, guiding LLMs to generate the final results.

**Table 10 | Structured JSON output formats for LLMs query responses**

| Method | JSON Response Formats |
|---|---|
| LLM-ZAVS (Local Query) | {"title": "answer", "type": "object", "description": "Output parsing structure.", "properties": {"reasoning": {"title": "Reasoning", "description": "The detailed reasoning behind the feature importance", "type": "string"}}, "required": ["reasoning"]} |
| LLM-ZAVS (Global Query) | {"title": "answer", "type": "object", "description": "Output parsing structure.", "properties": {"score and ranking": {"title": "score and ranking", "description": "feature score and ranking for predicting target variable", "type": "string"}, "reasoning": {"title": "Reasoning", "description": "The detailed reasoning behind the feature ranking.", "type": "string"}}, "required": ["score and ranking", "reasoning"]} |
| LLM-UFSS | {"title": "answer", "type": "object", "description": "Output parsing structure.", "properties": {"Reasoning": {"title": "Reasoning", "description": "The detailed reasoning behind the prediction result.", "type": "string"}, "Confidence Score": {"title": "Confidence Score", "description": "The confidence score behind the prediction result.", "type": "number"}, "Prediction Result": {"title": "Prediction result", "description": "Prediction result of target variable", "type": "number"}}, "required": ["Prediction Result", "Reasoning", "Confidence Score"]} |

To accurately extract model prediction results, confidence scores, and explanations from the LLM's output, we employ prompt engineering to enforce the model to return structured outputs in JSON format. This facilitates subsequent data processing. Instruction details are shown in Table 10.

# References

1. Yao L, Ge ZQ. Causal variable selection for industrial process quality prediction via attention-based GRU network. Eng Appl Artif Intell 118, 18 (2023).
2. Lawrence NP, et al. Machine learning for industrial sensing and control: A survey and practical perspective. Control Eng Practice 145, 16 (2024).
3. Zhang TM, Yan GW, Ren MF, Cheng L, Li R, Xie G. Dynamic transfer soft sensor for concept drift adaptation. J Process Control 123, 50-63 (2023).
4. Curreri F, Patanè L, Xibilia MG. Soft Sensor Transferability: A Survey. Appl Sci-Basel 11, 18 (2021).
5. Zhai RK, Zheng JH, Song ZH, Ge ZQ. Reliable Soft Sensors With an Inherent Process Graph Constraint. IEEE Trans Ind Inform 20, 8798-8806 (2024).
6. Zhang XM, He BC, Zhu HY, Song ZH. Information Complementary Fusion Stacked Autoencoders for Soft Sensor Applications in Multimode Industrial Processes. IEEE Trans Ind Inform 20, 106-116 (2024).
7. Niresi KF, Bissig H, Baumann H, Fink O. Physics-Enhanced Graph Neural Networks for Soft Sensing in Industrial Internet of Things. IEEE Internet Things J 11, 34978-34990 (2024).
8. Hasnen SH, Shahid M, Zabiri H, Taqvi SAA. Semi-supervised adaptive PLS soft-sensor with PCA-based drift correction method for online valuation of NOx emission in industrial water-tube boiler. Process Saf Environ Protect 172, 787-801 (2023).
9. Yuan XF, Ye LJ, Bao L, Ge ZQ, Song ZH. Nonlinear feature extraction for soft sensor modeling based on weighted probabilistic PCA. Chemometrics Intell Lab Syst 147, 167-175 (2015).
10. Wang ZX, He QP, Wang J. Comparison of variable selection methods for PLS-based soft sensor modeling. J Process Control 26, 56-72 (2015).
11. Li Z, Lee YS, Chen JH, Qian YW. Developing variable moving window PLS models: Using case of NOx emission prediction of coal-fired power plants. Fuel 296, 16 (2021).

12. Liu J, et al. A soft sensing method of billet surface temperature based on ILGSSA-LSSVM. Sci Rep 12, 10 (2022).

13. Wang HX, Sui L, Zhang MY, Zhang FF, Ma FY, Sun K. A Novel Input Variable Selection and Structure Optimization Algorithm for Multilayer Perceptron-Based Soft Sensors. Math Probl Eng 2021, 10 (2021).

14. Bendaouia A, et al. Artificial intelligence for enhanced flotation monitoring in the mining industry: A ConvLSTM-based approach. Comput Chem Eng 180, 13 (2024).

15. Yan F, Yang CJ, Zhang XM. Stacked Spatial-Temporal Autoencoder for Quality Prediction in Industrial Processes. IEEE Trans Ind Inform 19, 8625-8634 (2023).

16. Wang X, Liu H. Soft sensor based on stacked auto-encoder deep neural network for air preheater rotor deformation prediction. Adv Eng Inform 36, 112-119 (2018).

17. Jiang YC, Yin S, Dong JW, Kaynak O. A Review on Soft Sensors for Monitoring, Control, and Optimization of Industrial Processes. IEEE Sens J 21, 12868-12881 (2021).

18. Ge ZQ. Review on data-driven modeling and monitoring for plant-wide industrial processes. Chemometrics Intell Lab Syst 171, 16-25 (2017).

19. Yuan XF, et al. Attention-Based Interval Aided Networks for Data Modeling of Heterogeneous Sampling Sequences With Missing Values in Process Industry. IEEE Trans Ind Inform 20, 5253-5262 (2024).

20. Ma L, Wang MW, Peng KX. A missing manufacturing process data imputation framework for nonlinear dynamic soft sensor modeling and its application. Expert Syst Appl 237, 11 (2024).

21. Fortuna L, Graziani S, Xibilia MG. Comparison of Soft-Sensor Design Methods for Industrial Plants Using Small Data Sets. IEEE Trans Instrum Meas 58, 2444-2451 (2009).

22. Tian Y, Xu Y, Zhu QX, He YL. Novel Virtual Sample Generation Using Target-Relevant Autoencoder for Small Data-Based Soft Sensor. IEEE Trans Instrum Meas 70, 10 (2021).

23. Guo RY, Liu H, Xie G, Zhang YM, Liu D. A Self-Interpretable Soft Sensor Based on Deep Learning and Multiple Attention Mechanism: From Data Selection to Sensor Modeling. IEEE Trans Ind Inform 19, 6859-6871 (2023).

24. Kay H, et al. Constructing a Symbolic Regression-Based Interpretable Soft Sensor for Industrial Data Analytics and Product Quality Control. Ind Eng Chem Res 63, 4083-4092 (2024).

25. Rebello CM, et al. Machine Learning-Based Dynamic Modeling for Process Engineering Applications: A Guideline for Simulation and Prediction from Perceptron to Deep Learning. Processes 10, 18 (2022).

26. Achiam J, et al. Gpt-4 technical report. arXiv preprint arXiv:230308774, (2023).

27. Team G, et al. Gemini 1.5: Unlocking multimodal understanding across millions of tokens of context. arXiv preprint arXiv:240305530, (2024).

28. Zhao WX, et al. A survey of large language models. arXiv preprint arXiv:230318223, (2023).

29. Zhou JX, et al. Pre-trained multimodal large language model enhances dermatological diagnosis using SkinGPT-4. Nat Commun 15, 12 (2024).

30. Imani S, Du L, Shrivastava H. Mathprompter: Mathematical reasoning using large language models. arXiv preprint arXiv:230305398, (2023).

31. Gupta S, et al. PRISM: Patient Records Interpretation for Semantic clinical trial Matching system using large language models. npj Digit Med 7, 12 (2024).

32. Bran AM, Cox S, Schilter O, Baldassari C, White AD, Schwaller P. Augmenting large language models with chemistry tools. Nat Mach Intell 6, 13 (2024).

33. Wei J, et al. Emergent abilities of large language models. arXiv preprint arXiv:220607682, (2022).

34. Webb T, Holyoak KJ, Lu HJ. Emergent analogical reasoning in large language models. Nat Hum Behav 7, 1526-1541 (2023).

35. Dong Q, et al. A survey on in-context learning. In: Proceedings of the 2024 Conference on Empirical Methods in Natural Language Processing (2024).

36. Xie SM, Raghunathan A, Liang P, Ma T. An explanation of in-context learning as implicit bayesian inference. arXiv preprint arXiv:211102080, (2021).

37. Xie SW, Yu YJ, Xie YF, Tang ZH. Sensitive Feature Selection for Industrial Flotation Process Soft Sensor Based on Multiswarm PSO With Collaborative Search. IEEE Sens J 24, 17159-17168 (2024).

38. Lewis P, et al. Retrieval-augmented generation for knowledge-intensive nlp tasks. Advances in Neural Information Processing Systems 33, 9459-9474 (2020).

39. Gao Y, et al. Retrieval-augmented generation for large language models: A survey. arXiv preprint arXiv:231210997, (2023).

40. Wei J, et al. Chain-of-thought prompting elicits reasoning in large language models. Advances in neural information processing systems 35, 24824-24837 (2022).

41. Chu Z, et al. A survey of chain of thought reasoning: Advances, frontiers and future. arXiv preprint arXiv:230915402,  (2023).

42. Gao S, Wen X-C, Gao C, Wang W, Zhang H, Lyu MR. What makes good in-context demonstrations for code intelligence tasks with llms? In: 2023 38th IEEE/ACM International Conference on Automated Software Engineering (ASE)). IEEE (2023).

43. Shanahan M, McDonell K, Reynolds L. Role play with large language models. Nature 623, 493-498 (2023).

44. Li C, et al. Large language models understand and can be enhanced by emotional stimuli. arXiv preprint arXiv:230711760,  (2023).

45. Goldrick S, Stefan A, Lovett D, Montague G, Lennox B. The development of an industrial-scale fed-batch fermentation simulation. J Biotechnol 193, 70-82 (2015).

46. Munir N, et al. Interpretable Machine Learning Methods for Monitoring Polymer Degradation in Extrusion of Polylactic Acid. Polymers 15, 23 (2023).

47. Guo XP, Wang QQ, Li Y. Weighted target feature regression neural networks based soft sensing for industrial process. Can J Chem Eng 102, 840-852 (2024).

48. Li J, Huang H. Mutual Information Deep Sparse Auto-Encoding Hybrid DLSTM Prediction Network. Computer Engineering and Application 58, 277-285 (2022).

49. Zheng DD, Shao SM, Liu AN, Wang MS, Li T. Soft measurement model for wet gas flow rate based on ultrasonic and differential pressure sensing. Meas Sci Technol 35, 13 (2024).

50. Yang L, Liu H, Chen FG. Soft sensor method of multimode BOF steelmaking endpoint carbon content and temperature based on vMF-WSAE dynamic deep learning. High Temp Mater Process 42, 21 (2023).

51. Xu W, Tang J, Xia H, Yu W, Qiao JF. Multi-objective PSO semi-supervised random forest method for dioxin soft sensor. Eng Appl Artif Intell 135, 16 (2024).

52. Hua L, Zhang C, Sun W, Li YM, Xiong JL, Nazir MS. An evolutionary deep learning soft sensor model based on random forest feature selection technique for penicillin fermentation process. ISA Trans 136, 139-151 (2023).

53. Shin S, Baek K, Choi Y, So H. Utilization of Machine Learning Techniques in Hot-Film Based Airflow Rate Sensors for Improving Flow Measurement. Adv Intell Syst 6, 10 (2024).

54. Sadeghian A, Jan NM, Wu O, Huang B. Robust probabilistic principal component regression with switching mixture Gaussian noise for soft sensing. Chemometrics Intell Lab Syst 222, 11 (2022).

55. Chen N, Dai JY, Yuan XF, Gui WH, Ren WT, Koivo HN. Temperature Prediction Model for Roller Kiln by ALD-Based Double Locally Weighted Kernel Principal Component Regression. IEEE Trans Instrum Meas 67, 2001-2010 (2018).

56. Lee M, Bae J, Kim SB. Uncertainty-aware soft sensor using Bayesian recurrent neural networks. Adv Eng Inform 50, 14 (2021).

57. Guo RY, Liu H, Liu D. When Deep Learning-Based Soft Sensors Encounter Reliability Challenges: A Practical Knowledge-Guided Adversarial Attack and Its Defense. IEEE Trans Ind Inform 20, 2702-2714 (2024).

58. Guo RY, Chen QY, Liu H, Wang WQ. Adversarial Robustness Enhancement for Deep Learning-Based Soft Sensors: An Adversarial Training Strategy Using Historical Gradients and Domain Adaptation. Sensors 24, 15 (2024).

59. Chen L, Xu Y, Zhu QX, He YL. Adaptive Multi-Head Self-Attention Based supervised VAE for Industrial Soft Sensing With Missing Data. IEEE Trans Autom Sci Eng 21, 3564-3575 (2024).

60. Ren JY, Chen X, Zhao CH. Partial transfer learning network for data imputation and soft sensor under various operation conditions. J Cent South Univ 30, 3395-3413 (2023).

61. Ouyang L, et al. Training language models to follow instructions with human feedback. Advances in neural information processing systems 35, 27730-27744 (2022).